\documentclass[sigconf]{acmart}

\usepackage{booktabs} 
\usepackage{tabularx}
\usepackage{graphicx}
\usepackage{url}
\usepackage{mdwlist}
\usepackage{subcaption}
\usepackage{hyperref}
\hypersetup{colorlinks=false,linkcolor=blue,urlcolor=blue,citecolor=red}
\usepackage{amsfonts}
\usepackage[labelfont=bf,textfont=bf]{caption}
\usepackage{multirow}
\usepackage{array}
\usepackage[linesnumbered,ruled]{algorithm2e}
\SetKwComment{Comment}{$\triangleright$\ }{}
\usepackage{xcolor}
\usepackage{amssymb}
\usepackage[flushleft]{threeparttable}
\usepackage{enumitem}
\usepackage{balance}

\copyrightyear{2018} 
\acmYear{2018} 
\setcopyright{acmcopyright}
\acmConference[MM '18]{2018 ACM Multimedia Conference}{October 22--26, 2018}{Seoul, Republic of Korea}
\acmBooktitle{2018 ACM Multimedia Conference (MM '18), October 22--26, 2018, Seoul, Republic of Korea}
\acmPrice{15.00}
\acmDOI{10.1145/3240508.3240632}
\acmISBN{978-1-4503-5665-7/18/10}

\newcommand{\ie}{\textit{i.e.}}
\newcommand{\eg}{\textit{e.g.}}
\newcommand{\etal}{\textit{et al.}}

\begin{document}

\title{
Context-Aware Visual Policy Network \\ for Sequence-Level Image Captioning
}

\author{Daqing Liu}
\affiliation{\institution{University of Science and\\ Technology of China}}
\email{liudq@mail.ustc.edu.cn}

\author{Zheng-Jun Zha}
\authornote{Corresponding author.}
\affiliation{\institution{University of Science and\\ Technology of China}}
\email{zhazj@ustc.edu.cn}

\author{Hanwang Zhang}
\affiliation{\institution{Nanyang Technological University}}
\email{hanwangzhang@ntu.edu.sg}

\author{Yongdong Zhang}
\affiliation{\institution{University of Science and\\ Technology of China}}
\email{zhyd73@ustc.edu.cn}

\author{Feng Wu}
\affiliation{\institution{University of Science and\\ Technology of China}}
\email{fengwu@ustc.edu.cn}

\renewcommand{\shortauthors}{Liu et al.}

\begin{abstract}
Many vision-language tasks can be reduced to the problem of sequence prediction for natural language output. In particular, recent advances in image captioning use deep reinforcement learning (RL) to alleviate the ``exposure bias'' during training: ground-truth subsequence is exposed in every step prediction, which introduces bias in test when only predicted subsequence is seen. However, existing RL-based image captioning methods only focus on the language policy while not the visual policy (\eg, visual attention), and thus fail to capture the visual context that are crucial for compositional reasoning such as visual relationships (\eg, ``man riding horse'') and comparisons (\eg, ``smaller cat''). To fill the gap, we propose a Context-Aware Visual Policy network (CAVP) for sequence-level image captioning. At every time step, CAVP explicitly accounts for the previous visual attentions as the context, and then decides whether the context is helpful for the current word generation given the current visual attention. Compared against traditional visual attention that only fixes a single image region at every step, CAVP can attend to complex visual compositions over time. The whole image captioning model --- CAVP and its subsequent language policy network --- can be efficiently optimized end-to-end by using an actor-critic policy gradient method with respect to any caption evaluation metric. We demonstrate the effectiveness of CAVP by state-of-the-art performances on MS-COCO offline split and online server, using various metrics and sensible visualizations of qualitative visual context. The code is available at \url{https://github.com/daqingliu/CAVP}
\end{abstract}

%
%



\maketitle

\section{Introduction}

\begin{figure}
\centering
	\includegraphics[width=.9\linewidth]{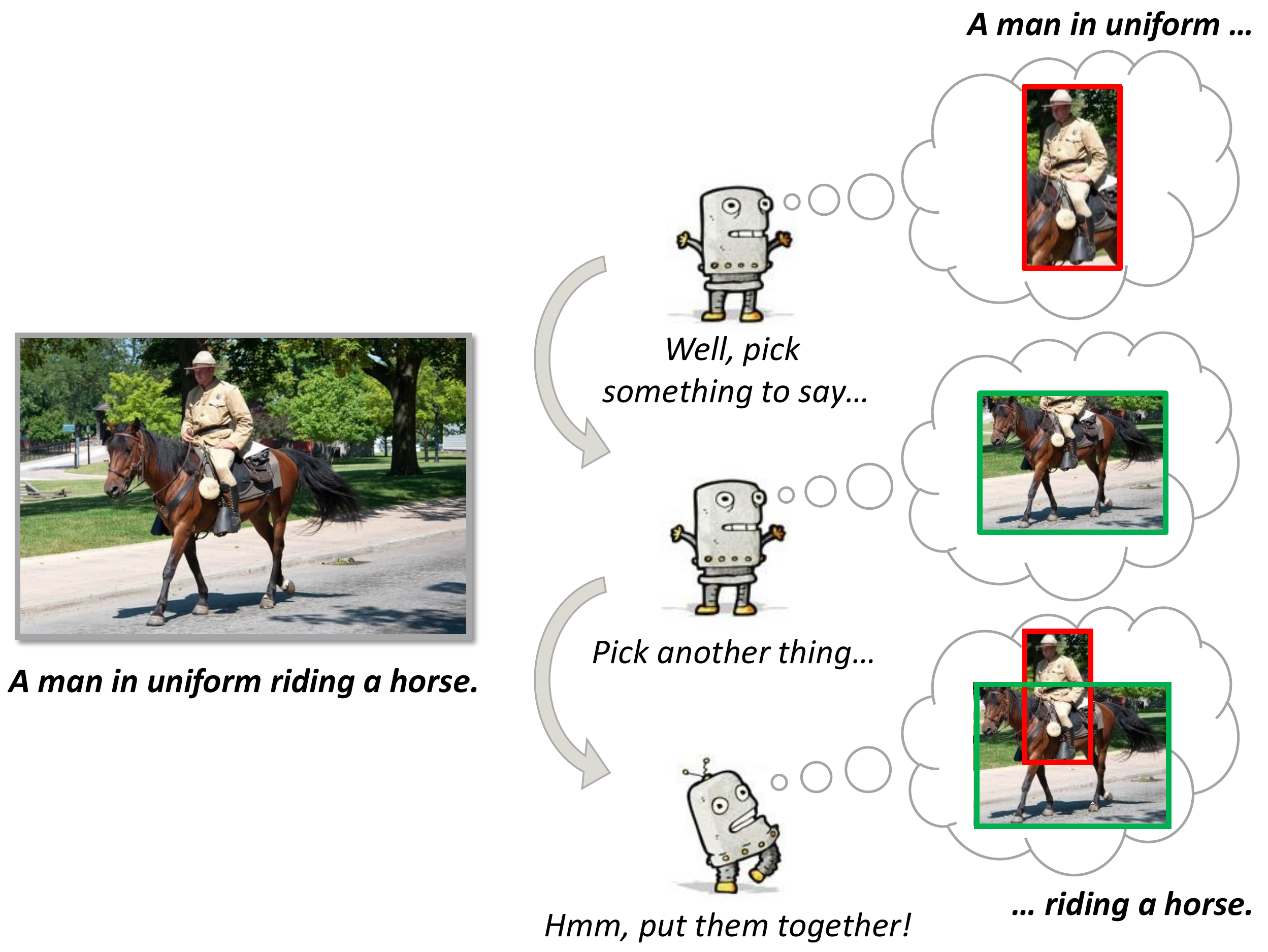}
    \vspace{-4mm}
    \caption{The intuition of using visual context in image captioning. We are the first RL-based image captioning model which incorporates visual context into sequential visual reasoning.}

    \label{fig:1}
\end{figure}

Visual and natural language comprehension --- the ever-lasting goal in the multimedia community --- are rapidly evolving with the help of deep learning based AI technologies~\cite{krishna2017visual,antol2015vqa,das2017visual}. A prime example is image captioning --- the task of generating natural language descriptions for an image, relying solely on the visual input --- which demonstrates a machine's visual comprehension in terms of its ability of natural language modeling~\cite{vinyals2015show,xu2015show}. As an AI-complete task~\cite{lake2017building}, researchers attempts to combine the most advanced computer vision (CV) techniques like object recognition~\cite{ren2015faster}, relationship detection~\cite{zhang2017visual}, and scene parsing~\cite{xu2017scene}, as well as the modern natural language processing (NLP) techniques such as language generative models~\cite{bahdanau2014neural,yu2017seqgan}. In a nutshell, the CV-end acts as an encoder and the NLP-end plays as an decoder, translating from ``source'' image to ``target'' sentences. This encoder-decoder architecture is trained using human-annotated image and sentence pairs in a fully-supervised way, that is, the decoder is supervised to maximize the posterior probability of each ground-truth word given the previous ground-truth subsequence and ``source'' image.  Unfortunately, due to the exponentially large search space of language compositions, recent studies demonstrate that this conventional supervised training tends to learn dataset bias but not machine reasoning~\cite{johnson2017inferring,jabri2016revisiting,hu2017learning}.

\begin{figure*}
	\begin{subfigure}{0.33\textwidth}
		\centering
		\includegraphics[width=\linewidth]{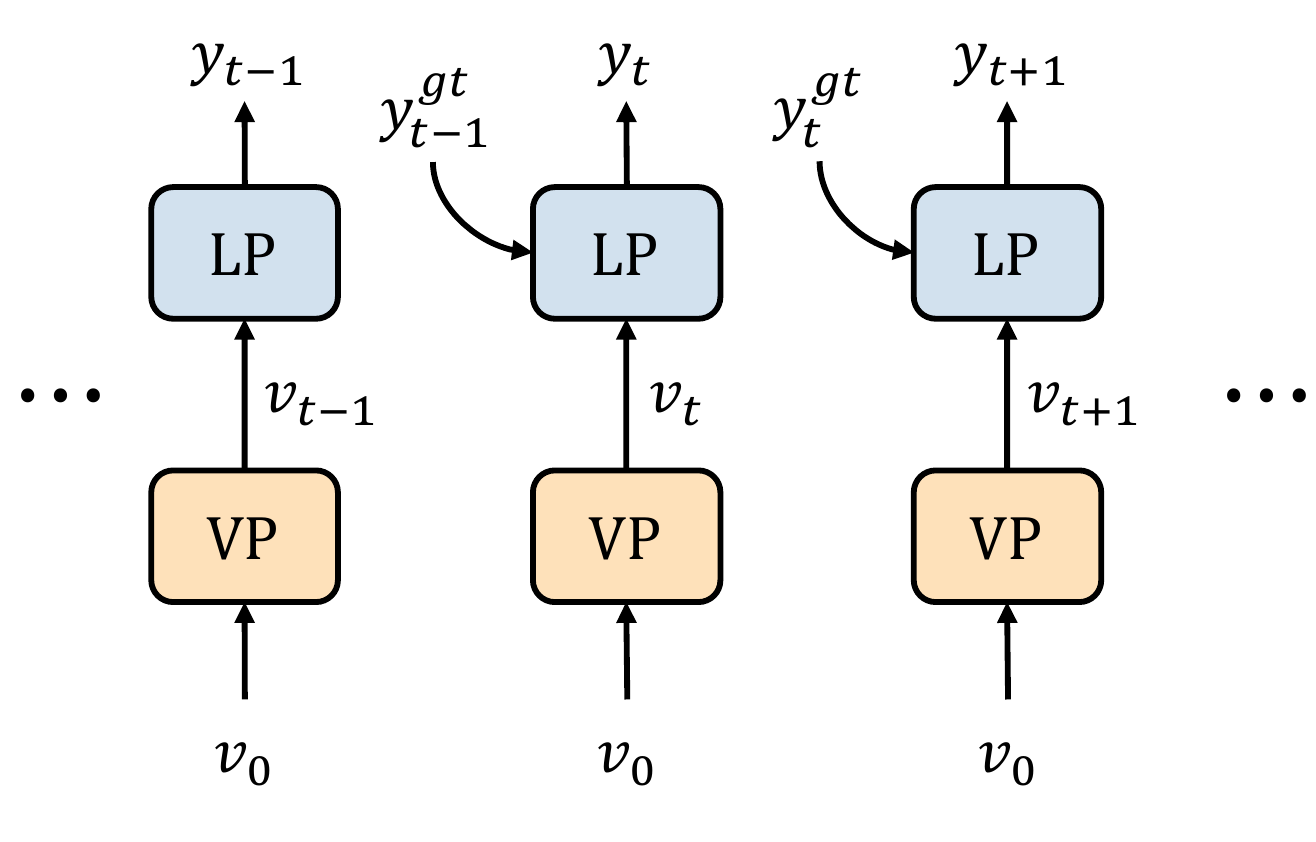}
		\caption{Traditional Framework}
        \label{fig:2a}
	\end{subfigure}
	\begin{subfigure}{0.33\textwidth}
		\centering
		\includegraphics[width=\linewidth]{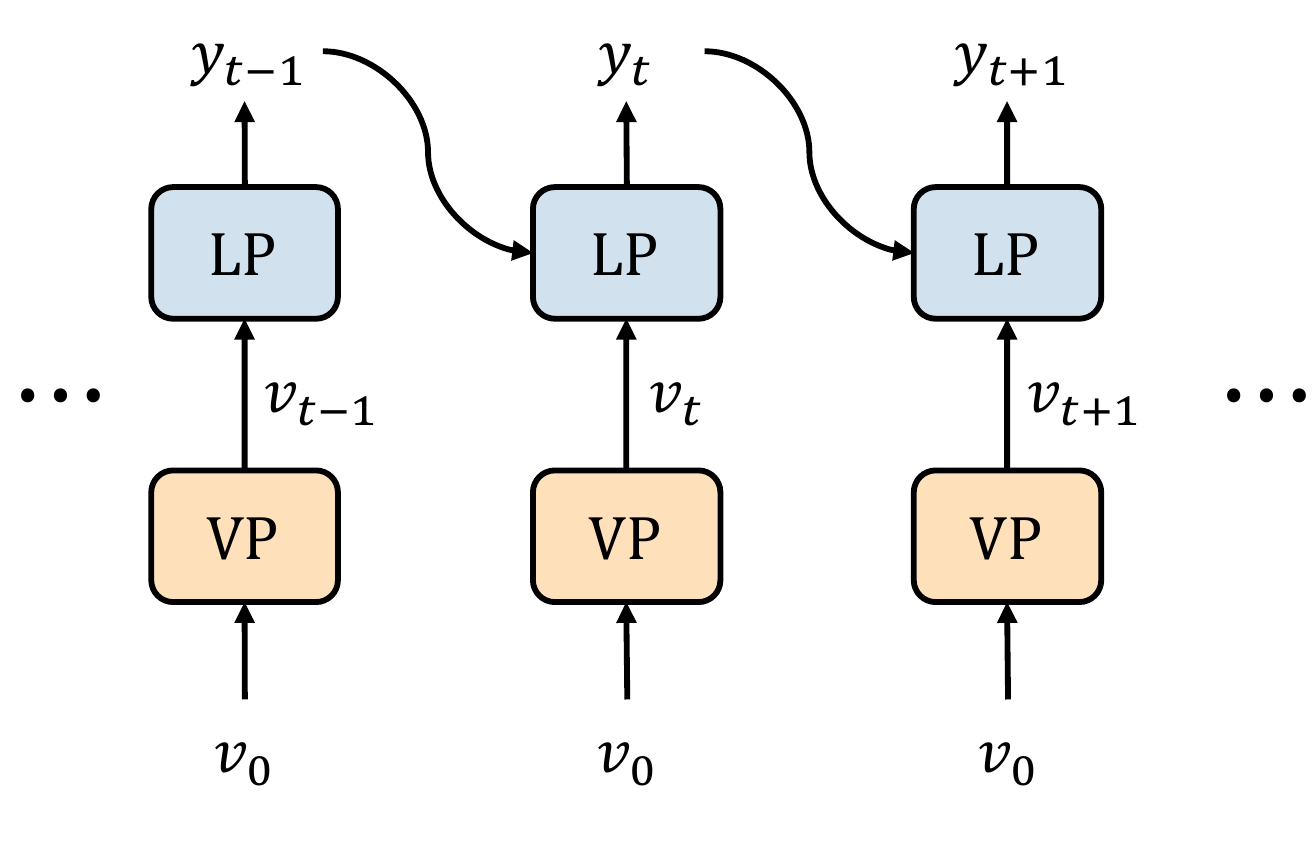}
		\caption{RL-based Framework}
        \label{fig:2b}
	\end{subfigure}
	\begin{subfigure}{0.33\textwidth}
		\centering
		\includegraphics[width=\linewidth]{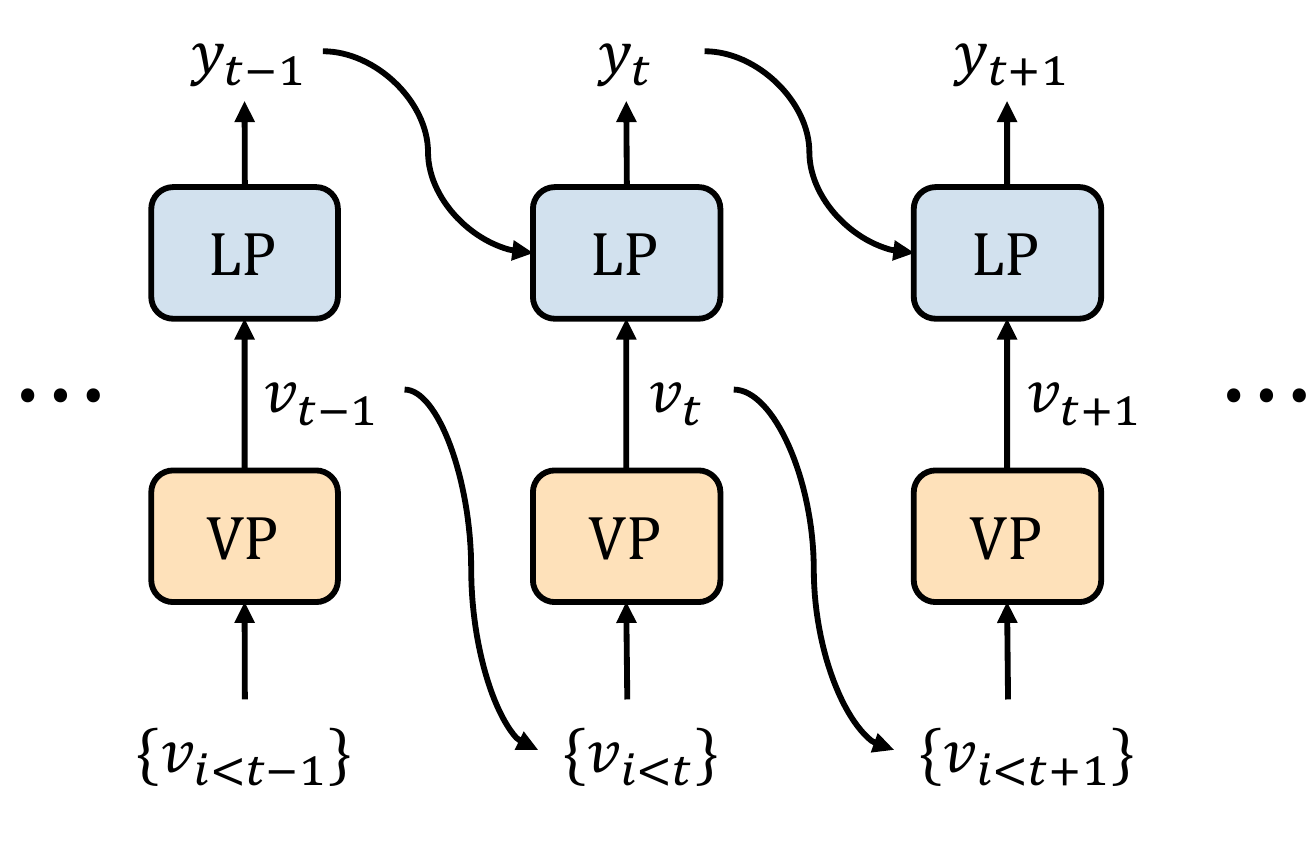}
		\caption{Our Framework}
        \label{fig:2c}
	\end{subfigure}
	\caption{The evolution of the encoder-decoder framework for image captioning. LP: language policy. VP: visual policy. $v_t$: visual feature at time $t$. $y_t$: predicted word at time $t$. $y^{gt}_t$: ground-truth word at time $t$. (a) traditional framework only focuses on the word-level prediction by exposing the ground-truth word $y^{gt}_{t-1}$ as input to step $t$ for sentence generation. (b) RL-based framework focuses on sequence-level training by directly input the predicted word $y_{t-1}$ to LP at step $t$. (c) our framework explicitly takes the history visual actions $\{v_{i < t}\}$ as visual context to step $t$.}
\vspace{-4mm}
	\label{fig:2}
\end{figure*}

An emerging line of training strategy is the sequence-level training using deep reinforcement learning (RL)~\cite{liu2016improved,rennie2016self,ren2017deep,zhang2017actor}. To see this, as illustrated in Figure~\ref{fig:2a}, we first frame the traditional encoder-decoder image captioning into the decision-making process, where the visual encoder can be viewed as Visual Policy (VP): deciding where to fix a gaze in the image, and the language decoder can be viewed as Language Policy (LP): deciding what the next word is. As highlighted in Figure~\ref{fig:2b}, the sequence-level RL-based framework directly injects the previously sampled word (sampling but not greedy argmax) to influence the next prediction. This brings two benefits: 1) training supervision is delayed to the whole sequence generated, thus we can use non-differentiable sequence-level metric such as CIDEr~\cite{vedantam2015cider} and SPICE~\cite{anderson2016spice}, which are more suitable than word-level cross entropy loss for language quality evaluation; 2) it avoids the exposure bias~\cite{ranzato2015sequence}, \ie, the LP has never been exposed to the ground-truth at training, allowing exploration over the large sequence compositions under current policy, and thus generating more fruitful sentences without severe overfitting. 

However, existing RL-based framework neglects to turn VP into decision-making, \eg, the input of VP is identical in every step as shown in Figure~\ref{fig:2b}. This disrespects the nature of sequence prediction, where the history visual actions (\eg, previously attended regions) should explicitly influence the current visual policy. To this end, we develop a Context-Aware Visual Policy (CAVP) network for sequence-level image captioning. As shown in Figure~\ref{fig:2c}, CAVP allows the previous visual features, \ie, the previous output of CAVP, to serve as the visual context for the current action. Different from the conventional visual attention~\cite{xu2015show}, where the visual context is \emph{implicitly} encoded in a hidden RNN state vector from LP, our visual context is \emph{explicitly} considered in a sequence prediction process. Our motivation is in line with the cognitive evidences that the visual memory recall plays a crucial role in compositional reasoning~\cite{stanfill1986toward}. For example, as illustrated in Figure~\ref{fig:1}, it is necessary to consider the contextual regions, \eg, the previously selected ``man'' object, when generating the composition ``man riding a horse''.

We decompose CAVP into 4 sub-policy networks, which together accomplish the visual decision-making task (cf. Figure~\ref{fig:3}), each of which is an Recurrent Neural Network (RNN) control with shared Long Short-Term Memory (LSTM) parameters and outputs a soft visual attention map. As we will show in Section~\ref{sec:visual_policy}, this CAVP design stabilizes the conventional Monte Carlo policy rollout and reduces the exponentially large search complexity to linear time. It is worth noting that CAVP and its subsequent language policy network can efficiently model higher-order compositions over time, \eg, relationships among multiple objects mentioned in the generated sub-sequence. The whole framework is trained end-to-end using an actor-critic policy gradient with a self-critic baseline~\cite{rennie2016self}. In fact, our CAVP can be seamlessly integrated into any policy-based RL models~\cite{sutton1998reinforcement}. We show the effectiveness of the proposed CAVP through extensive experiments on the MS-COCO image captioning benchmark~\cite{lin2014microsoft}. By using only a single model, we can beat all the published RL-based methods on MS-COCO server. In particular, we significantly improve every SPICE~\cite{anderson2016spice} compositional scores such as object, relation, and attribute without optimizing on it. We also show promising qualitative results of visual policy reasoning over the time of generation.

\section{Related Work}
\subsection{Image Captioning}
Inspired by the recent advances in machine translation \cite{bahdanau2014neural}, current image captioning approaches \cite{vinyals2015show, xu2015show, lu2017knowing, anderson2017bottom, chen2017sca, chen2017structcap} typically follow an encoder-decoder framework, which can be considered as a neural translation task from image to text. It uses CNN-RNN architectures that encodes an image as feature vectors by CNN~\cite{krizhevsky2012imagenet, he2016deep} and decodes such vectors to a sentence by RNN~\cite{hochreiter1997long}.

More recently, attention mechanisms which allows dynamic feature vectors have been introduced to the encoder-decoder framework. Xu \textit{et al.}~\cite{xu2015show} incorporated \textit{soft} and \textit{hard} attention mechanism to automatically focus on salient objects when generating corresponding words. Chen \etal ~\cite{chen2017sca} introduced channel-wise attention besides spatial attention mechanism and Lu \textit{et al.}~\cite{lu2017knowing} proposed a visual sentinel to deal with the non-visual words when generating captions. Besides the spatial information comes from CNN feature maps, Anderson \textit{et al.}~\cite{anderson2017bottom} used an object detection network to propose salient image regions with an associated feature vector as bottom-up attention mechanism.
However, all those captioning approaches only focus on the current time step's visual attention and neglect to consider the visual context over time, which is the key for language compositions. To this end, we introduce the context-aware visual policy network which incorporates history visual attentions to current time step and generate more effective feature vectors fed into word policy network.
\subsection{Sequential Decision-Making}
Most recent captioning approaches are typically trained by maximizing the likelihood of the ground-truth word at each time steps. This causes the exposure bias~\cite{ranzato2015sequence} between the training and the testing phases. To mitigate it, reinforcement learning have been applied to image captioning which introduces the notion of sequential decision-making.
The idea of making a series of decisions forces the agent to take into account future sequences of actions, states, and rewards.
In the case of image captioning, the state is the image, previously generated words and visual context, the action is choosing next word and visual representation, and the reward can be any metric of interest.

Several attempts have been made to apply sequential decision-making framework to image captioning.
The first work~\cite{ranzato2015sequence} proposed by Ranzato \etal~trained RNN-based sequence model by policy gradient method based on a Monte Carlo search technique, where policy gradient was used to optimize the sentence-level reward.
Rennie \etal~\cite{rennie2016self} used the classic REINFORCE algorithm~\cite{williams1992simple} and applied a relative baseline obtained by the current model to the final reward. In result, for each sampled caption, it has a sentence level advantage indicating how good or bad this sentence is. It assume every token makes the same contribution towards the whole sentence.
Actor-Critic based method ~\cite{zhang2017actor} was also applied to image captioning by utilized two networks as Actor and Critic respectively.
After that, Ren \etal~\cite{ren2017deep} introduced decision-making framework utilized a policy network and a value network with embedding reward to collaboratively generate captions.

In our work, we formulate the image captioning task into a sequence-level training framework where each word-level prediction policy is based on the action performed by the proposed CAVP. Our framework is optimized using policy gradient with a self-critic value which can directly optimize non-differentiable quality metrics of interest such as CIDEr \cite{vedantam2015cider}.

\section{Approach}
\begin{figure*}
	\begin{center}
		\includegraphics[width=\linewidth]{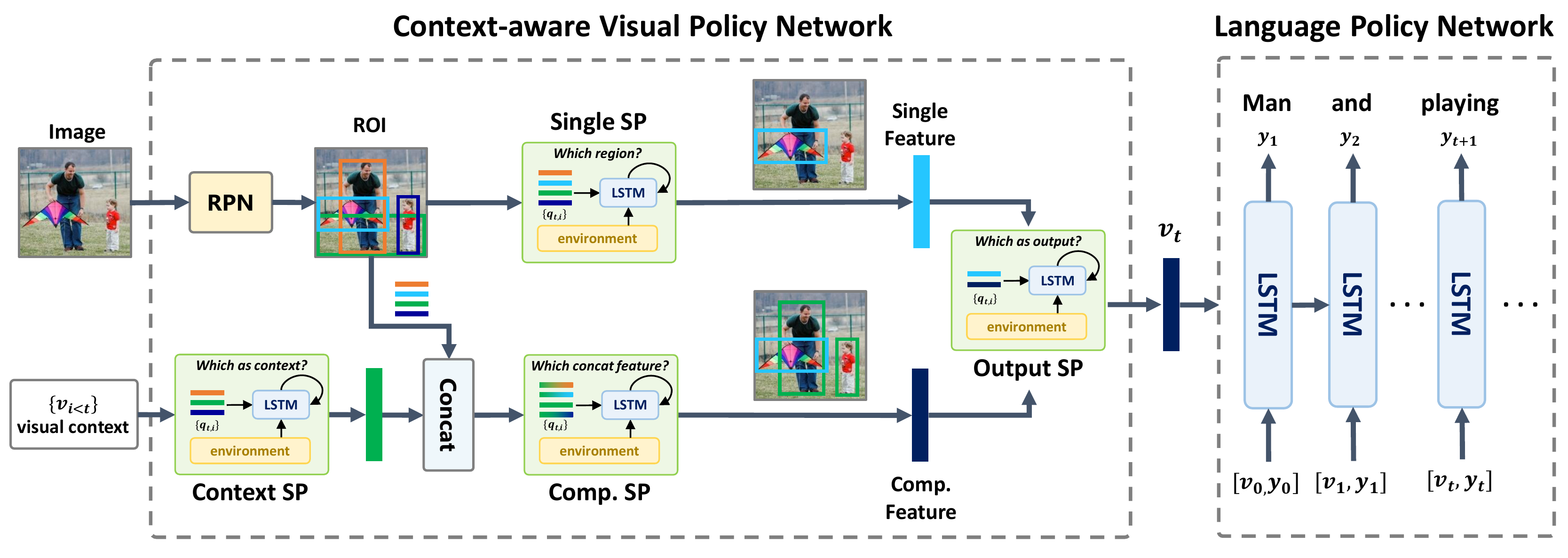}
	\end{center}
	\caption{Overview of the proposed RL-based image captioning framework. It consists of the proposed CAVP for visual feature composition and the language policy for sentence generation. CAVP contains 4 sub-policy (SP) networks: Single SP (Section~\ref{sec:single}),  Context SP (Section~\ref{sec:context}), Composition SP (Section~\ref{sec:composition}), and Output SP (Section~\ref{sec:output}). The notations used include visual context: $\{v_{i<t}\}$, query inputs: $\{q_{t,i}\}$, output visual feature: $v_t$, and predicted words: $y_t$.}
\label{fig:3}
\end{figure*}


In this section, we explain our approach in more detail. The overview of the proposed RL-based image captioning framework is illustrated in Figure~\ref{fig:3}.
We first define our problem formulation, then we detail the Context-Aware Visual Policy network (CAVP) and the language policy network. Finally we discuss the sequence-level training strategy for the entire framework.

\subsection{Problem Formulation}
We formulate the task of image captioning into a sequential decision-making process.
In decision-making, there is an \emph{agent} interacts with the \emph{environment}, and then executes a series of \emph{actions}, after each \emph{action}, a \emph{state} is observed, so as to optimize the \emph{reward} when accomplishing a \emph{goal}.

In image captioning, given an image $\mathbf{I}$, the \emph{goal} is to generate a sequential visual representations $\{v_1, v_2, \cdots, v_T\}$, where $v_t$ is the $t$-th action of visual policy network, and a corresponding sentence $Y=\{y_1, y_2, \cdots, y_T\}, y_t \in \Sigma$, where $\Sigma$ is the vocabulary, $y_t$ is the $t$-th action of language policy network, \ie, generated word.

Our model, including the visual policy network and the language policy network, can be viewed as the \emph{agent}. At time step $t$, the \emph{state} includes the given image $\mathbf{I}$, visual context $\{v_{i<t}\}$ and the generated sequence so far $\{y_{i<t}\}$. An action is to predict the next visual representation $v_t$ and word $y_t$. The \emph{reward} is any sequence evaluation between the ground-truth $Y^{gt}$ and the generated $Y$.

For image representation, we use Faster R-CNN~\cite{ren2015faster} to extract image region features $\{r_1, \cdots, r_k\}$ from image $\mathbf{I}$. We select top $k$ ROIs ranked according to their ROI scores in each image. For every selected region $i$, $r_i$ is the mean-pooling convolutional feature from this region.

\subsection{Context-Aware Visual Policy Network}
\label{sec:visual_policy}
The proposed CAVP is designed to select the most informative image features for the subsequent language policy network (LP). A brute force search of all possible contextual regions requires $\mathcal{O}(2^N)$ complexity for multinomial combinations of $N$ image regions. For linear efficiency, we approximate the overall visual policy network into 4 sub-policy networks: 1) single sub-policy network $\pi^s$, 2) composition sub-policy network $\pi^p$, 3) context sub-policy network $\pi^c$, and 4) output sub-policy network $\pi^o$.

Each sub-policy network uses a RNN with the LSTM cell~\cite{hochreiter1997long} to encode the environment state. Moreover, we use soft-attention features as the real-valued action performed by the sub-policy networks, which can be considered as the deterministic approximation for the Monte Carlo policy search~\cite{zhang2018grounding}, to reduce the sample variance caused by the diverse image regions. Here, without loss of generality, we first introduce the general structure of the sub-policy networks denoted as $\pi$ without any superscripts.

At time $t$, a sub-policy network as an \emph{agent} receives environment \emph{state} $s_t$, then take an \emph{action} $a_t \sim \pi(s_t|h_t;\theta)$, where $h_t$ is it's LSTM hidden state.
Then the sub-policy network according to this \emph{action} sampling a representation $f$ from query inputs. Query inputs $Q_t=\{q_{t,1}, q_{t,2}, \cdots , q_{t,d}\}$ are a set of features which are difference depending on specific sub-policy network and we will determine it in the next section. The general formulation is given by:
\begin{equation}
f = \sum_{i=1}^d \pi(a_t=i) q_{t,i},
\end{equation}
with the LSTM hidden state transition function:
\begin{equation}
h_{t+1} = \mathrm{LSTM}(s_t, h_t).
\end{equation}
To compute the probability of taking action $a_t$, we follow the additive attention mechanism~\cite{xu2015show} as:
\begin{equation}\label{eqn:softmax}
\pi(a_t=i) = \mathrm{softmax}(w_a^T \tan(W_h h_t + W_q q_{t,i})),
\end{equation}
where $\pi(a_t=i) \in [0, 1]$, and $w_a$, $W_h$ and $W_q$ are trainable parameters.

In this way, if we determine the environment state $s_t$ and query inputs $Q_t$ at each time steps of a sub-policy network, the sequence decision-making is known. Next, we will implement each sub-policy networks by introducing the corresponding state and query inputs.

\subsubsection{Single Sub-policy Network}
\label{sec:single}
For single sub-policy network, the environment state at each time steps consist of the previous LSTM hidden state $h_{t-1}^l$ of the language policy network, concatenated with the mean-pooled region features $\bar{r} = \frac{1}{k} \sum_{i=1}^k r_i$, and word embedding of the previously generated word:
\begin{equation}
s_t^s = [h_{t-1}^l, \bar{r}, W_e\Pi_{t-1}],
\end{equation}
where $W_e \in \mathbb{R}^{E\times\Sigma}$ is a word embedding matrix of a vocabulary size $\Sigma$, which is learned from random initialization, and $\Pi_{t-1}$ is one-hot encoding of the output action (\ie\ generated word) of the language policy network at time step $t-1$.
The query inputs at each time steps are the detected region features, \ie $Q_t^s = \{r_1, r_2, \cdots, r_k\}$.
We take the output of single sub-policy network as single feature at time $t$: $v_t^s\leftarrow f_t^s$. Single feature will be used in output sub-policy network (Section \ref{sec:output}).

\subsubsection{Context Sub-policy Network}
\label{sec:context}
At time step $t$, we have the visual contexts $\{v_{i<t}\}$; however, not all contexts are helpful to generate the current time step's word. Therefore, we introduce the context sub-policy network to choose the most informative context and combine it with the detected region features.
In detail, we define the environment state as:
\begin{equation}
s_t^c = [h_{t-1}^l, \bar{r}, W_e\Pi_{t-1}],
\end{equation}
and the query inputs as $Q_t^c = \{v_{i<t}\}$.

Then we fuse visual context representation $f_t^c$ at step $t$, \ie\ the output from context sub-policy network, with every region features to concatenate features $c_{t,i}$ as:
\begin{equation}
c_{t,i} = W_c^T[f_t^c; r_i], \  i = 1, 2, \cdots, k,
\end{equation}
where $[\cdot;\cdot]$ indicates concatenating two vectors and $W_c^T$ projecting the concatenated features to the original dimension as region features. Concatenate features will be used in composition sub-policy network (Section~\ref{sec:composition}).
Besides, we can also use a simple but effective way that only considering the visual context from the last time step $t-1$, \ie:
\begin{equation}
c_{t, i} = W_c^T[v_{t-1}, r_i], \  i = 1, 2, \cdots, k,
\end{equation}
which will reduce the computation without sacrificing performance. We will discuss this approximation in Section~\ref{sec:ablation}.

\subsubsection{Composition Sub-policy Network}
\label{sec:composition}
The composition sub-policy network is similar to the single sub-policy network which takes the previous hidden state of the language policy network, the mean-pooled region features, and an encoding of the previously generated word as environment state:
\begin{equation}
s_t^p = [h_{t-1}^l, \bar{r}, W_e\Pi_{t-1}].
\end{equation}
The query inputs of the composition sub-policy network takes concatenate features from the context sub-policy network:
\begin{equation}
Q_t^p = \{c_{t,1}, c_{t,2}, \cdots, c_{t,k}\}.
\end{equation}
Then we take the output of composition sub-policy network as composition features at time $t$: $v_t^p\leftarrow f_t^p$.

\subsubsection{Output Sub-policy Network}
\label{sec:output}
After obtaining the single and compositional visual features from Single SP and Comp. SP\@, we given the environment state of output sub-policy network by:
\begin{equation}
s_t^o = [h_{t-1}^l, \bar{r}, W_e\Pi_{t-1}].
\end{equation}
And determined query inputs as $Q_t^o = \{v_t^s, v_t^p\}$.

The output of this sub-policy is thus the overall CAVP's output visual representation, \ie\ $v_t = f_t^o$, which will be used in language policy network at each time steps and also will be seen as a part of visual context in subsequent time steps.

\subsubsection{Weights Sharing}
We noticed that although the above sub-policy network are difference in query inputs, but some of them have the same environment state like:
\begin{equation}
s_t^c = s_t^s = s_t^p = s_t^o = [h_{t-1}^l, \bar{r}, W_e\Pi_{t-1}].
\end{equation}
To reduce the complexity of our model and alleviate over-fittings, we shared the LSTM parameters among those sub-policy networks in experiments. More ablative studies of the weight sharing is detailed in Section~\ref{sec:ablation}.

\subsection{Language Policy Network}
\label{sec:LP}
At each time step, CAVP generates a context-aware visual representation most fitting to current words. Language policy network take the visual representation and the hidden state $h_t^s$ of Single SP as input, then use them to update LSTM hidden state:
\begin{equation}
h_{t+1}^l = \mathrm{LSTM}([h_t^s, v_t], h_t^l).
\end{equation}
To compute the distribution over all words in vocabulary, we apply a FC layer to hidden state, and after softmax layer it outputs the probability distribution of each words, given by:
\begin{equation}\label{eqn:y_t}
\pi^l(y_t | y_{1:t-1}) = \mathrm{softmax} (W_y h^l_t + b_y),
\end{equation}
where $W_y$ and $b_y$ are learnable weights and biases. For complete output sentences, the distribution is calculated as the product of all time step's conditional distributions:
\begin{equation}\label{math_output}
\pi^l(y_{1:T})=\prod_{t=1}^T \pi^l(y_t|y_{1:t-1}).
\end{equation}

\subsection{Sequence-Level Training}
For sequence-level training, we use two learning stages: 1) standard supervised learning with cross entropy loss and 2) reinforcement learning by policy gradient method using a self-critical relative base reward~\cite{rennie2016self}.
As described in the previous section, traditional captioning models are trained using the cross entropy loss.
Given a target ground-truth sequence $y_{1:T}^{gt}$ and a captioning model with parameters $\theta$, the traditional supervised learning approach is used to train this network by minimizing the cross entropy loss (XE):
\begin{equation}\label{eqn:xe_loss}
L_{XE}(\theta) = -\sum_{t=1}^{T}\log(\pi_{\theta}(y_t^{gt} \mid y_{1:t-1}^{gt})).
\end{equation}
This corresponds to the imitation learning of a perfect teacher in RL context, and we use the pre-trained model as the initial policy network.

To directly optimize the NLP metrics and address the exposure bias issue, we use the policy gradient method to maximize the expected reward, for instance, the CIDEr score of the generated sequences. Initializing from the cross-entropy trained model, we minimize the negative expected score:
\begin{equation}
L_{R}(\theta) = -E_{y \sim \pi_\theta}{r(y_{1:T})},
\end{equation}
where $r$ is the metric evaluation (\eg~BCMR). Following the~\cite{rennie2016self} approach, the gradient of this loss can be approximated to:
\begin{equation}\label{eqn:rl_gradient}
\nabla_{\theta}L_{R}(\theta) \approx -(r(y_{1:T}^s)-r(\hat{y}_{1:T}))\nabla_{\theta}\log \pi_\theta(y_{1:T}^s),
\end{equation}
where $y_{1:T}^s$ is a caption sampled according to the word distribution (\ie~Monte-Carlo sampling) and $\hat{y}_{1:T}$ is a greedy searched caption.  The REINFORCE~\cite{williams1992simple} algorithms explores the space of captions by sampling captions from the policy network. While training, this gradient tends to increase the probability of each words in the sampled captions if $r(y_{1:T}^s)$ higher than $r(\hat{y}_{1:T})$, which can been seen as the relative baseline score, and vice versa.

\subsubsection{Behavior Cloning}
The learning would be easier if we have some additional knowledge of the output policy.
While there no any additional knowledge in the caption datasets e.g. MS-COCO, we can use a language parser \cite{toutanova2003feature} as an existing expert output policy that can be used to provide additional supervision.
More generally, if there is an expert output policy $\pi^e$ that predicts a reasonable output policy $\pi^o$, we can first pre-train our model by behavioral cloning from $\pi^e$. This can be done by minimizing the KL-divergence $D_{KL}(\pi^e || \pi^o)$ between the expert output policy $\pi^e$ and our output policy $\pi^o$, and simultaneously minimizing the captioning loss $L_{XE}$ with expert output policy $\pi^e$. This supervised behavioral cloning from the expert output policy can provide a good set of initial parameters in our output sub-policy network.
Note that the above behavioral cloning procedure is only done at cross-entropy training time to obtain a supervised initialization for our model, and the expert output policy is not used at test time.

The expert output policy is not necessarily optimal, for behavioral cloning itself is not sufficient for learning the most suitable output policy for each image.
After learning a good initialization by cloning the expert output policy, our model is further trained end-to-end with gradient $\nabla_{\theta}L_{R}(\theta)$ computed using Eqn.~\eqref{eqn:rl_gradient}, where the output policy $\pi^o$ is sampled from the output policy network in our model, and the expert output policy $\pi^e$ can be discarded.

\section{Experiments}
In this section, we first describe the datasets used in our experiments and the widely used evaluation metrics. Then, we go through the implementation details and the comparing methods. Finally, we analyze both of the quantitative and qualitative results.

\subsection{Dataset}
We used the most popular benchmark \textbf{MS-COCO}~\cite{lin2014microsoft} for image captioning, which contains 82,783 images for training and 40,504 for validation. Each image is human-annotated with 5 captions. As the annotations of the official test set are not publicly available, for validation of model hyperparameters and offline testing, we follow the widely used ``Karpathy'' splits~\cite{karpathy2015deep} in most prior works, containing 113,287 images for training, 5,000 for validation, and 5,000 for testing. We reported the results both on ``Karpathy'' offline split and MS-COCO online test server.

\subsection{Metrics}
The most common metrics for caption evaluation are based on $n$-gram similarity of reference and candidate descriptions.
\textbf{BLEU}~\cite{papineni2002bleu} is defined as the geometric mean of n-gram precision scores, with a sentence-brevity penalty.
In \textbf{CIDEr}~\cite{vedantam2015cider}, $n$-grams in the candidate and reference sentences are weighted by term frequency-inverse document frequency weights (\ie~tf-idf). Then, the cosine similarity between them are computed.
\textbf{METEOR}~\cite{banerjee2005meteor} is defined as the harmonic mean of precision and recall of exact, stem, synonym, and paraphrase matches between sentences.
\textbf{ROUGE}~\cite{lin2004rouge} is a measures for automatic evaluation for summarization systems via F-measures.
We followed~\cite{ren2017deep} to call them them \textbf{BCMR} and used all of them as metrics.

All the above metrics are originally developed for the evaluation of text summaries or machine translations. It has been shown that there are obvious bias between those metrics and human judgment~\cite{anderson2016spice}. Therefore, we further evaluated our model using \textbf{SPICE}~\cite{anderson2016spice} metric, which is defined over tuples that are divided into semantically meaningful categories such as objects, relations and attributes.

\subsection{Implementation Details}
\subsubsection{Data Pre-processing}
We performed standard minimal text pre-processing: first tokenizing on white space, second converting all words into lower case, then filtering out words that occur less than 5 times, finally resulting in a vocabulary of 10,369 words. Captions are trimmed to a maximum of 16 words for computational efficiency.
To generate a set of image region features $r_i$ in image captioning, we take the final output of the region proposed network~\cite{ren2015faster} and perform non-maximum suppression. In our implementation we used an IoU threshold of 0.7 for region proposal non-maximum suppression, and 0.3 for object class non-maximum suppression. To select salient image regions, we simply selected the top $k=36$ features in each image for computation consider.

\subsubsection{Parameter Settings}
We set the number of hidden units of each LSTM to 1,300, the number of LSTM layers to 1, the number of hidden units in the attention-like mechanism we described in Eqn.~\eqref{eqn:softmax} to 1,024, and the size of word embedding to 1000.
During the supervised learning for the cross-entropy process, we used Adam optimizer~\cite{kingma2014adam} with base learning rate of 5e-4 and shrink it by 0.8 every 3 epochs. We start reinforcement learning after 37 epochs, we used Adam optimizer with base learning rate of 5e-5 and shrink it by 0.1 every 55 epochs.
We set the batch size to 100 images and trained for up to 100 epochs. During inferencing stage, we use a beam search size of 5.
While training Faster R-CNN, we follow~\cite{anderson2017bottom} and first initialize it with ResNet-101~\cite{he2016deep} pretrained with classification on ImageNet~\cite{russakovsky2015imagenet}, then fine-tune it on Visual Genome~\cite{krishna2017visual} with attribute labels.

\subsection{Comparing Methods}
\subsubsection{Traditional Framework}
We first compared our models with classic methods including~\textbf{Google NIC}~\cite{vinyals2015show}, \textbf{Hard Attention}~\cite{xu2015show}, \textbf{Adaptive Attention}~\cite{lu2017knowing} and \textbf{LSTM-A}~\cite{yao2016boosting}. These methods follow the popular encoder-decoder architecture, trained with cross-entropy loss between the predicted and ground-truth words, that is, no sequence-level training is applied.
\subsubsection{RL-based Framework}
We also compared our models with the RL-based methods including \textbf{PG-SPIDEr-TAG}~\cite{liu2016improved}, \textbf{SCST}~\cite{rennie2016self}, \textbf{Embedding-Reward}\cite{ren2017deep}, and \textbf{Actor-Critic}~\cite{zhang2017actor}. These methods use sequence-level training with various reward returns such as the metrics mentioned above.

\subsubsection{Ablative Models}
\label{sec:ablation}
We extensively investigated ablative structures and settings of our model to gain insights about how and why it works:
1) \textbf{Single}: we only used the single sub-policy network as the visual policy network without context.
2) $\mathbf{CAVP_{4c}}$: we shared the LSTM weights among all \textbf{4} sub-policy networks (\ie context, single, compositional and output sub-policy networks) with fully \textbf{c}ontext-aware. Attention that we only shared LSTM parameters and not all parameters in a sub-policy network.
3) $\mathbf{CAVP_{4p}}$: we shared LSTM weights among all \textbf{4} sub-policy networks (\ie, context, single, compositional and output sub-policy networks) and simply used the \textbf{p}revious visual feature as visual context.
4) $\mathbf{CAVP_{3p+scratch}}$: we shared LSTM weights among \textbf{3} sub-policy networks (\ie context, single, and composition sub-policy networks) and simply used the \textbf{p}revious visual feature as visual context. And while training output sub-policy network, we did not use expert policy and trained from \textbf{scratch}.
5) $\mathbf{CAVP_{3p+cloning}}$: we shared the weights among \textbf{3} sub-policy networks (\ie context, single and composition sub-policy networks) and simply used the \textbf{p}revious visual feature as visual context. And the output sub-policy network is \textbf{cloning} from expert policy.

\begin{table*}[!t]
    \begin{center}\small
        \setlength{\tabcolsep}{.35em}
        \begin{tabular}{lccccccccccccccccccccc}
            \midrule
            & \multicolumn{2}{c}{BLEU-1} &  & \multicolumn{2}{c}{BLEU-2} &  & \multicolumn{2}{c}{BLEU-3} &  & \multicolumn{2}{c}{BLEU-4} &  & \multicolumn{2}{c}{METEOR} &  & \multicolumn{2}{c}{ROUGE-L} &  & \multicolumn{2}{c}{CIDEr} \\
            \cmidrule{2-3}\cmidrule{5-6}\cmidrule{8-9}\cmidrule{11-12}\cmidrule{14-15}\cmidrule{17-18}\cmidrule{20-21}
            & c5           & c40         &  & c5           & c40         &  & c5           & c40         &  & c5           & c40         &  & c5           & c40         &  & c5           & c40         &  & c5           & c40        \\
            \midrule
            Google NIC\cite{vinyals2015show} &71.3&89.5& &54.2&80.2& &40.7&69.4& &30.9&58.7& &25.4&34.6& &53.0&68.2& &94.3&94.6\\
            MSR Captivator\cite{fang2015captions}&71.5&90.7& &54.3&81.9& &40.7&71.0& &30.8&60.1& &24.8&33.9& &52.6&68.0& &93.1&93.7\\
            M-RNN\cite{mao2014deep}&71.6&89.0& &54.5&79.8& &40.4&68.7& &29.9&57.5& &24.2&32.5& &52.1&66.6& &91.7&93.5\\
            Hard-Attention\cite{xu2015show} &70.5&88.1& &52.8&77.9& &38.3&65.8& &27.7&53.7& &24.1&32.2& &51.6&65.4& &86.5&89.3\\
        Adaptive\cite{lu2017knowing} &74.8&92.0& &58.4&84.5& &44.4&74.4& &33.6&63.7& &26.4&35.9& &55.0&70.5& &104.2&105.9\\
            PG-SPIDEr-TAG\cite{liu2016improved}&75.1&91.6& &59.1&84.2& &44.5&73.8& &33.6&63.7& &25.5&33.9& &55.1&69.4& &104.2&107.1\\
            SCST:Att2all\cite{rennie2016self}&78.1&93.7& &61.9&86.0& &47.0&75.9& &35.2&64.5& &27.0&35.5& &56.3&70.7& &114.7&116.7\\
            LSTM-A$_{\text{3}}$ \cite{yao2016boosting}&78.7&93.7& &62.7&86.7& &47.6&76.5& &35.6&65.2& &27&35.4& &56.4&70.5& &116.0&118.0\\
            Stack-Cap\cite{gu2017stack}&77.8&93.2& &61.6&86.1& &46.8&76.0& &34.9&64.6& &27.0&35.6& &56.2&70.6& &114.8&118.3\\
            Up-Down\cite{anderson2017bottom}& \textbf{80.2} & \textbf{95.2} &  & 64.1 & 88.8 &  & 49.1 & 79.4 &  & 36.9 & 68.5 &  & 27.6 & 36.7 &  & 57.1 & 72.4 &  & 117.9 & 120.5\\
            \midrule
            Ours & 80.1 & 94.9 &  & \textbf{64.7} & \textbf{88.8} &  & \textbf{50.0} & \textbf{79.7} &  & \textbf{37.9} & \textbf{69.0} &  & \textbf{28.1} & \textbf{37.0} &  & \textbf{58.2} & \textbf{73.1} &  & \textbf{121.6} & \textbf{123.8}\\
            \midrule
        \end{tabular}
    \end{center}
    \caption{Highest ranking published image captioning results on the online MSCOCO test server. Except for BLUE-1 which is of little interest, our single model optimized with CIDEr, outperforms previously published works using all the other metrics.}
    \vspace{-3mm}
    \label{tab:online}
\end{table*}

\subsection{Comparisons with State-of-The-Art}
\subsubsection{Quantitative Analysis}
\begin{figure}
\centering
\includegraphics[width=0.9\linewidth]{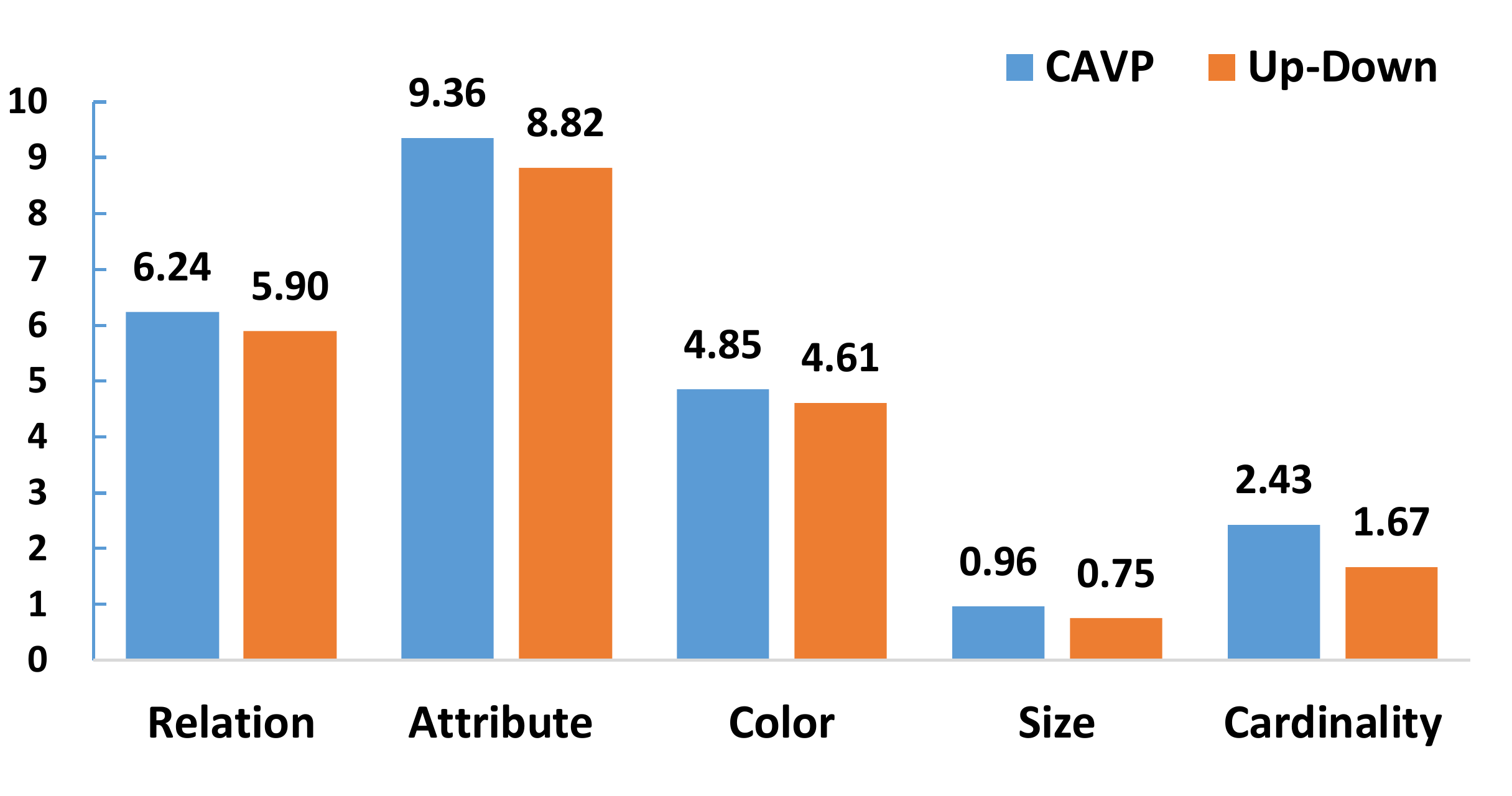}
\caption{The performance of our $\mathbf{CAVP_{4p}}$ model and Up-Down model which is very similar to our Single policy model. All SPICE category scores are improved by CAVP.}
\label{fig:spice}
\end{figure}

\begin{table}[]
	\centering
	\begin{tabular}{l|ccccc}
		\hline
		Model          & B@4  & M    & R    & C     & S    \\ \hline
		Google NIC\cite{vinyals2015show}   & 32.1 & 25.7 & -    & 99.8  & -    \\
		Hard-Attention\cite{xu2015show} & 24.3 & 23.9 & -    & -     & -    \\
		Adaptive\cite{lu2017knowing}       & 33.2 & 26.6 & 54.9 & 108.5 & 19.4 \\
		LSTM-A\cite{yao2016boosting}         & 32.5 & 25.1 & 53.8 & 98.6  & -    \\ \hline
		PG-SPIDEr\cite{liu2016improved}      & 32.2 & 25.1 & 54.4 & 100.0 & -    \\
		Actor-Critic\cite{zhang2017actor}  & 34.4 & 26.7 & 55.8 & 116.2 & - \\
        EmbeddingReward\cite{ren2017deep}           & 30.4 & 25.1 & 52.5 & 93.7 & -    \\
		SCST\cite{rennie2016self}           & 35.4 & 27.1 & 56.6 & 117.5 & -    \\
		StackCap\cite{gu2017stack}       & 36.1 & 27.4 & 56.9 & 120.4 & 20.9 \\
		Up-Down\cite{anderson2017bottom}        & 36.3 & 27.7 & 56.9 & 120.1 & 21.4 \\ \hline
		Ours           & \textbf{38.6} & \textbf{28.3} & \textbf{58.5} & \textbf{126.3} & \textbf{21.6} \\ \hline
	\end{tabular}
	\caption{
		Performance comparisons on MS-COCO ``Karpathy'' offline split. B@n is short for BLEU-n, M is short for METEOR, R is short for ROUGE, C is short for CIDEr, and S is short for SPICE.
	}
\vspace{-8mm}
    \label{tab:offline}
\end{table}
As shown in Table \ref{tab:offline}, we evaluated our model compared with the state-of-the-art methods.
We found that almost all RL-based methods outperform traditional methods.
The reason is that RL addresses the loss-evaluation mismatch problem and included the inference process in training to address the exposure bias problem.
We can also find that our CAVP outperforms other non-context methods. This is because the visual context information is useful for current word generation and the policy makes better decisions. In particular, we achieved state-of-the-art performance under all metrics on ``Karpathy'' test split.
Besides the BCMR metrics, our SPICE scores also outperform all the other methods. SPICE correlates with human judgment better and can be detailed into semantic categories. Our $\mathbf{CAVP_{4p}}$ model improves all category scores compared with the Up-down\cite{anderson2017bottom} model shown in Figure~\ref{fig:spice}.

\subsubsection{Qualitative Analysis.}

\begin{figure}\label{fig:word}
	\centering
	\begin{subfigure}{0.4\linewidth}
    	\centering
		\includegraphics[height=\linewidth]{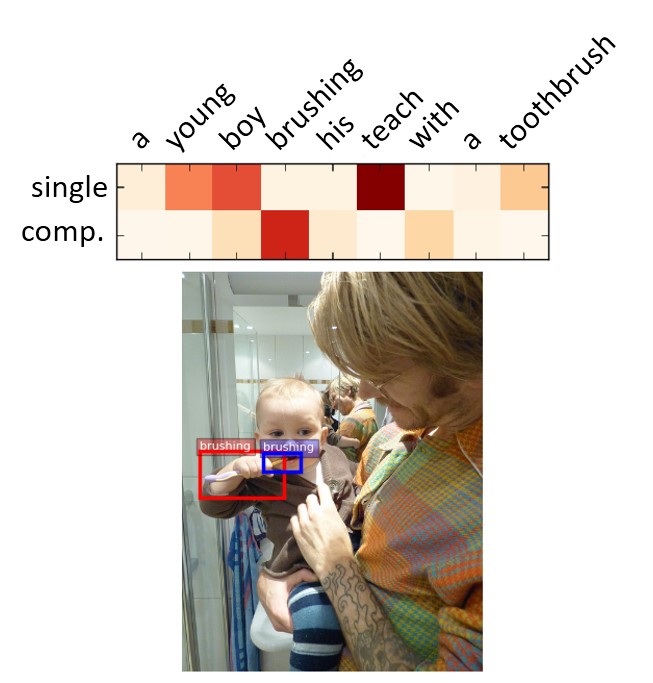}
		\caption{brushing}
        \label{fig:word_brushing}
	\end{subfigure}%
	\begin{subfigure}{0.4\linewidth}
        \centering
		\includegraphics[height=\linewidth]{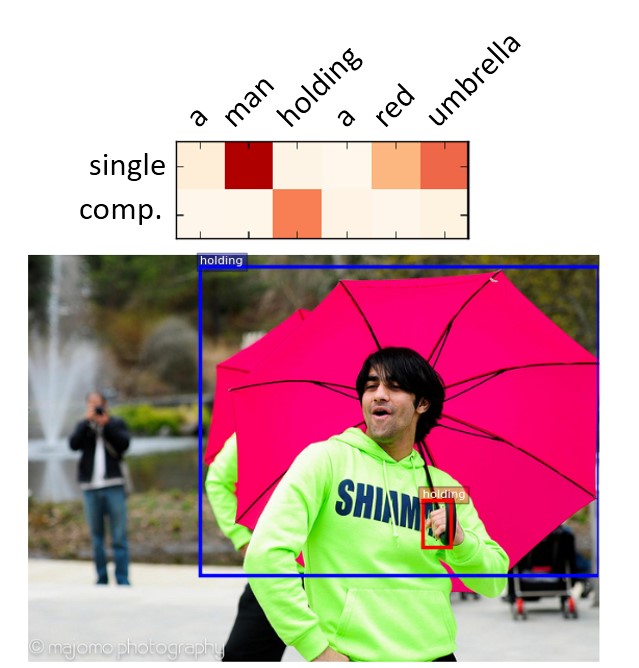}
		\caption{holding}
        \label{fig:word_holding}
	\end{subfigure}
    
	\begin{subfigure}{0.4\linewidth}
		\centering
		\includegraphics[height=\linewidth]{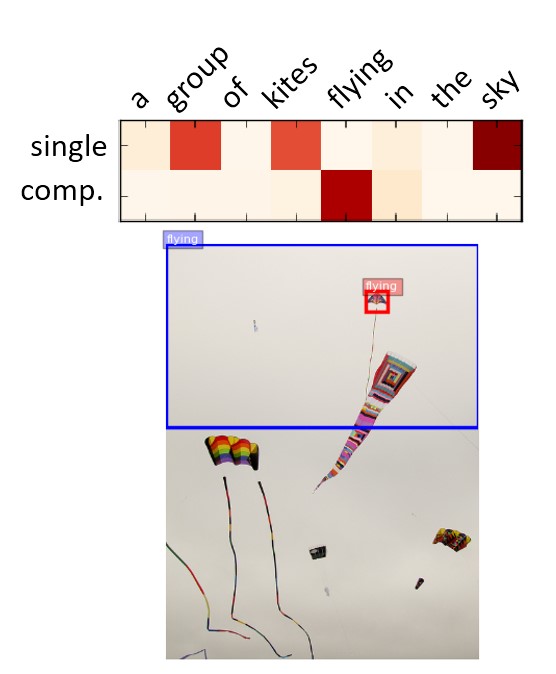}
		\caption{flying}
        \label{fig:word_flying}
	\end{subfigure}%
	\begin{subfigure}{0.4\linewidth}
		\centering
		\includegraphics[height=\linewidth]{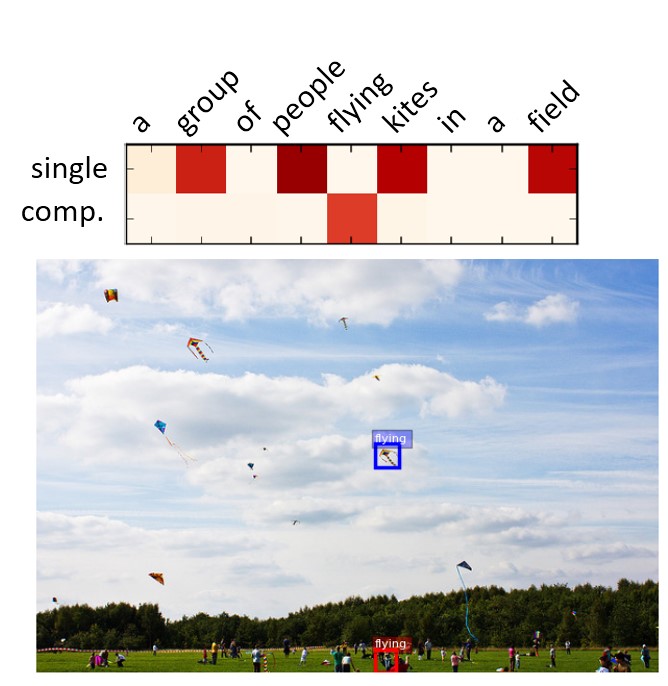}
		\caption{flying}
        \label{fig:word_flying2}
	\end{subfigure}
	\caption{
    Qualitative examples where top matrix shows the output policy network action probabilities and the bottom image shows the decision with maximum probability for composition features. The blue bounding boxes are the context regions and the red bounding boxes are the current regions which concatenated with context regions.
}
\vspace{-4mm}
	\label{fig:word}
\end{figure}

To better understand our CAVP, we show some qualitative visualizations as well as the output sub-policy network's predictions in Figure ~\ref{fig:word}.
Take Figure~\ref{fig:word_brushing} as an example, after we generated ``a young boy'', we first focus on the visual context, the boy's hand which is holding something, and then we want to find visual regions that the boy is holding, so we focus on the toothbrush in the boy's mouth. Note that both hand and toothbrush we generated is exactly the word ``brushing''. In Figure ~\ref{fig:word_flying} and ~\ref{fig:word_flying2}, although they generate the same words, the context of these two words are different from Figure~\ref{fig:word_flying} where the context is ``kite''; while in Figure~\ref{fig:word_flying2}, the context is ``people'' when we want to caption what the people are doing. By applying our CAVP, we can generate those captions both successfully with deep understanding of image scenes.
Besides showing a single important word of the generated sequence, we also visualize the whole policy decision across the whole sentence generation in Figure~\ref{fig:sent}.
Take the first sentence as an example, we notice that our context-aware model can not only focus on some single objects such as ``man'', ``skis'', and ``snow'', but also the compositional word ``standing'', connecting ``man standing in snow''.
\begin{figure*}\label{fig:sent}
	\begin{subfigure}{.48\linewidth}
	\centering
	\includegraphics[width=\linewidth]{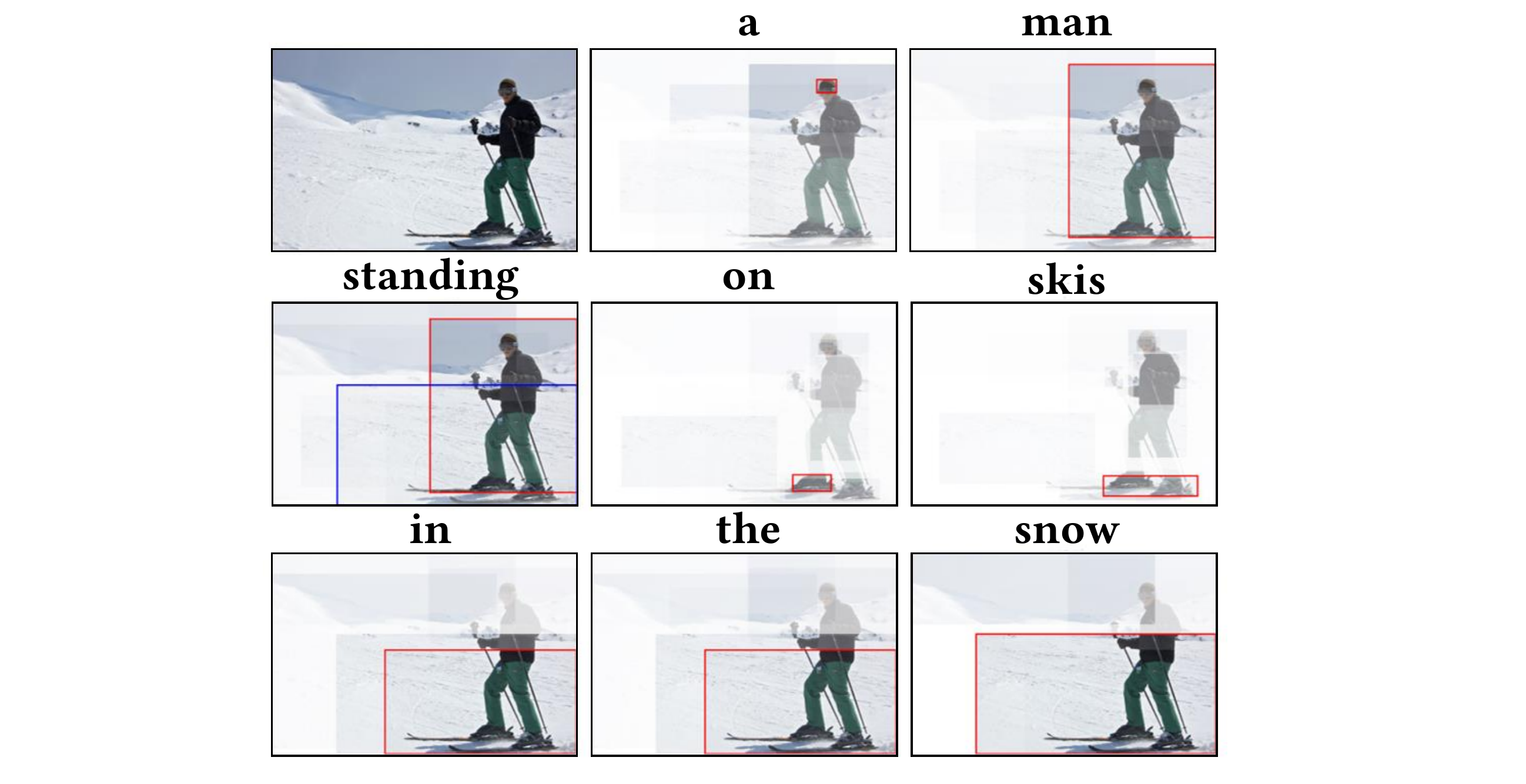}
	\caption{a man standing on skis in the snow}
    \end{subfigure}
\begin{subfigure}{.48\linewidth}
	\centering
	\includegraphics[width=\linewidth]{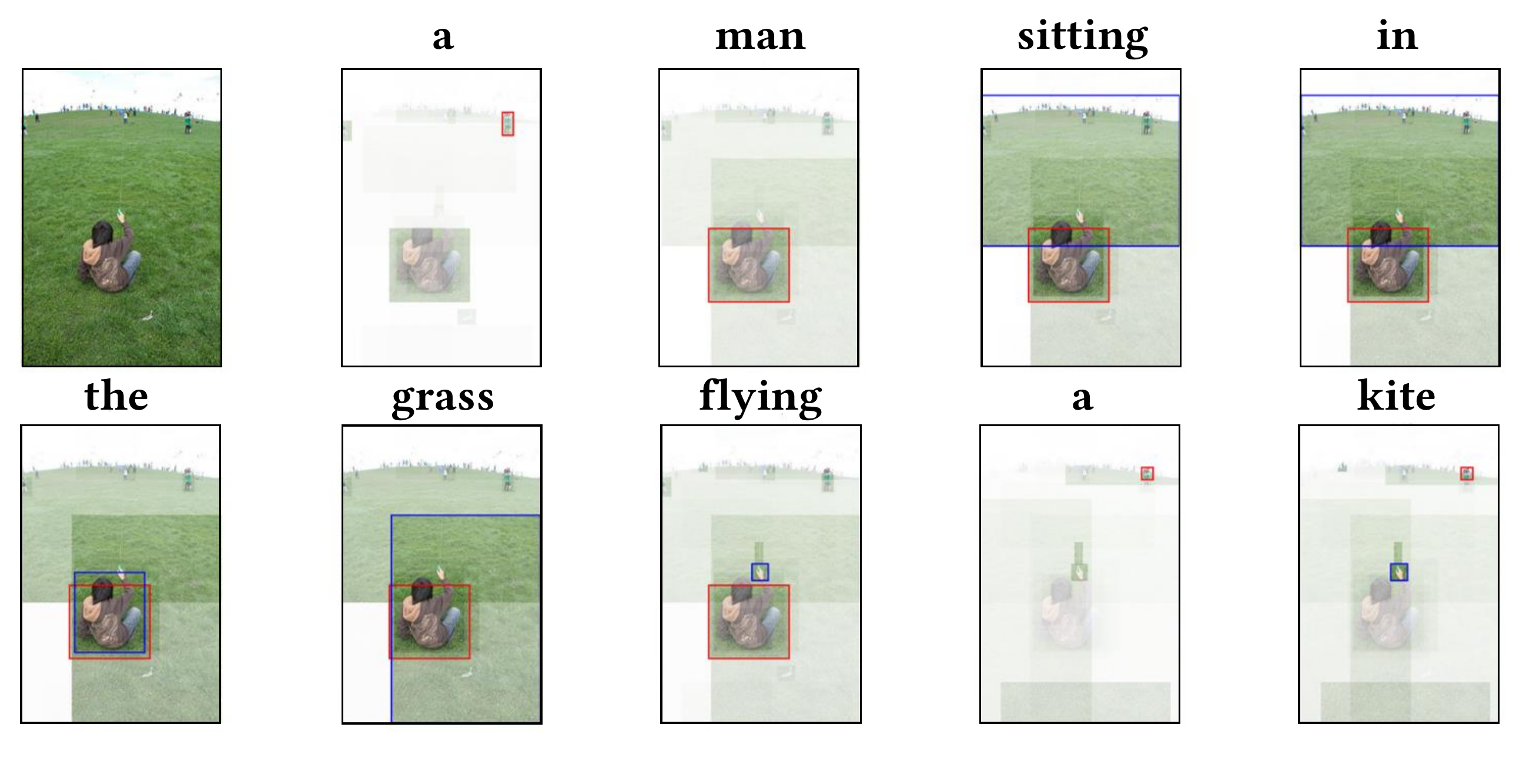}
	\caption{a man sitting in the grass flying a kite}
\end{subfigure}
\caption{For each generated word, we visualized the attended image regions, outlining the region with the maximum policy probability in bounding box. The blue bounding boxes are the visual context representation regions and the red bounding boxes are the regions decided by single policy network.}
\label{fig:sent}
\end{figure*}

\subsubsection{Online Results.}
Table~\ref{tab:online} reports the performances of our single model without any ensemble on the official MS-COCO evaluation server\footnote[1]{\url{https://competitions.codalab.org/competitions/3221\#results}}. Note that the highest results are from industrial companies while we are academic groups who lack of abundant computing sources. Moreover, we only used a single model while theirs are fused models.
With more computing resource for parameter fine-tuning, we believe that there will be a large space for improving our models.

\subsection{Ablative Studies}
\subsubsection{Architecture}
Under the framework there are a lot of variants, we do some ablative studies for insights into our framework.
As shown in Table~\ref{tab:ablation}, we found that sharing the weights between all the 4 sub-policy networks gains the best result. It is because without weight sharing, the model is easily overfitted.
We also found that $\mathbf{CAVP_{4p}}$ is similar to the $\mathbf{CAVP_{4c}}$ which indicates that the most informative visual context is from the last time step and little comes from older ones.
Comparing the $\mathbf{CAVP_{3p+scrach}}$ model and $\mathbf{CAVP_{3p+cloning}}$ model, with the expert policy guiding, there is little improvement. This indicates that the off-the-sheaf language parser is not very suitable to the visual-language task and the output sub-policy network can learned from scratch without any initialization.
\begin{table}[]
\centering
\begin{tabular}{l|ccccc}
\hline
Model & B@4 & M & R & C & S \\ \hline
\textbf{Single} & 37.5 & 27.7 & 57.9 & 121.9 & 21.0 \\
$\mathbf{CAVP_{3p+scrach}}$ & 37.8 & 28.0 & 58.2 & 124.5 & 21.3 \\
$\mathbf{CAVP_{3p+cloning}}$ & 38.3 & 27.8 & 58.0 & 124.6 & 21.4 \\
$\mathbf{CAVP_{4p}}$ & 38.3 & 28.2 & 58.4 & \textbf{126.4} & 21.6 \\
$\mathbf{CAVP_{4c}}$ & \textbf{38.6} & \textbf{28.3} & \textbf{58.5} & 126.3 & \textbf{21.6} \\ \hline
\end{tabular}
\caption{Ablative performance comparisons on MS-COCO. B@n is short for BLEU-n, M is short for METEOR, R is short for ROUGE, C is short for CIDEr, and S is short for SPICE.}
\vspace{-6mm}
\label{tab:ablation}
\end{table}
\subsubsection{Reward}
\begin{table}[]
\centering
\begin{tabular}{c|ccccc}
\hline
\multirow{2}{*}{\begin{tabular}[c]{@{}c@{}}Training\\ Metric\end{tabular}} & \multicolumn{5}{c}{Evaluation Metric} \\
 & BLEU4 & ROUGE & METEOR & CIDEr & SPICE \\ \hline
BLEU & \textbf{38.8} & 57.7 & 27.3 & 114.5 & 20.7 \\
ROUGE & 38.1 & \textbf{59.1} & 27.8 & 120.0 & 20.8 \\
METEOR & 33.6 & 57.6 & \textbf{29.6} & 113.0 & 22.8 \\
CIDEr & 38.3 & 58.4 & 28.2 & \textbf{126.4} & 21.6 \\
SPIDEr & 37.8 & 58.0 & 27.8 & 125.3 & \textbf{23.1} \\ \hline
\end{tabular}
\caption{
Ablative performances on the MS-COCO ``Karpathy'' offline split with respect to various metrics as the reward.
}
\vspace{-8mm}
\label{tab:reward}
\end{table}
For sequence-level training by policy gradient, the reward function $r$ can be any metrics, \eg~BLEU, ROUGE, METEOR, CIDEr and SPIDEr~\cite{liu2016improved} (which combining the CIDEr and SPICE scores equally as the reward). Optimizing for different metrics leads to different performance. In general, as shown in Table~\ref{tab:reward}, we found that optimizing for a specific metric results in the best performance on the same metric. And optimizing for CIDEr and SPIDEr gives the best overall performance, but the SPIDEr is more time consuming as the SPICE metric evaluation is very slow. Thus, we chose the CIDEr as the optimizing objective in most our experiments.

\section{Conclusion}
In this paper, we presented a sequence-level image captioning framework using a novel Context-Aware Visual Policy network (CAVP).
Unlike existing RL-based methods, our framework takes the advantage of visual context in compositional visual reasoning, which is beneficial for image captioning. Compared against traditional visual attention which only fixes a single image region at every step, CAVP can attend to complex visual compositions over time. To the best of our knowledge, we are the first RL-based image captioning model which incorporates visual context into sequential visual reasoning.
We conducted extensive experiments as well as ablative studies to show the effectiveness of CAVP. Our framework can significantly boost the performances of the RL-based image captioning methods and achieves top ranking performances on MS-COCO server. Moving forward, we are going to 1) integrate the visual policy and language policy into a Monte Carlo tree search strategy for sentence generation, and 2) apply CAVP in other sequential decision-making tasks such as visual Q\&A and visual dialog.

\section*{Acknowledgments}
This work was supported by the National Key R\&D Program of China under Grant 2017YFB1300201, the National Natural Science Foundation of China
(NSFC) under Grants 61622211, 61472392 and 61620106009 as well as the 
Fundamental Research Funds for the Central Universities under Grant WK2100100030.

\bibliographystyle{ACM-Reference-Format}
\balance
\bibliography{citation}


\begin{thebibliography}{44}


\ifx \showCODEN    \undefined \def \showCODEN     #1{\unskip}     \fi
\ifx \showDOI      \undefined \def \showDOI       #1{#1}\fi
\ifx \showISBNx    \undefined \def \showISBNx     #1{\unskip}     \fi
\ifx \showISBNxiii \undefined \def \showISBNxiii  #1{\unskip}     \fi
\ifx \showISSN     \undefined \def \showISSN      #1{\unskip}     \fi
\ifx \showLCCN     \undefined \def \showLCCN      #1{\unskip}     \fi
\ifx \shownote     \undefined \def \shownote      #1{#1}          \fi
\ifx \showarticletitle \undefined \def \showarticletitle #1{#1}   \fi
\ifx \showURL      \undefined \def \showURL       {\relax}        \fi
\providecommand\bibfield[2]{#2}
\providecommand\bibinfo[2]{#2}
\providecommand\natexlab[1]{#1}
\providecommand\showeprint[2][]{arXiv:#2}

\bibitem[\protect\citeauthoryear{Anderson, Fernando, Johnson, and
  Gould}{Anderson et~al\mbox{.}}{2016}]%
        {anderson2016spice}
\bibfield{author}{\bibinfo{person}{Peter Anderson}, \bibinfo{person}{Basura
  Fernando}, \bibinfo{person}{Mark Johnson}, {and} \bibinfo{person}{Stephen
  Gould}.} \bibinfo{year}{2016}\natexlab{}.
\newblock \showarticletitle{Spice: Semantic propositional image caption
  evaluation}. In \bibinfo{booktitle}{\emph{ECCV}}.
\newblock


\bibitem[\protect\citeauthoryear{Anderson, He, Buehler, Teney, Johnson, Gould,
  and Zhang}{Anderson et~al\mbox{.}}{2017}]%
        {anderson2017bottom}
\bibfield{author}{\bibinfo{person}{Peter Anderson}, \bibinfo{person}{Xiaodong
  He}, \bibinfo{person}{Chris Buehler}, \bibinfo{person}{Damien Teney},
  \bibinfo{person}{Mark Johnson}, \bibinfo{person}{Stephen Gould}, {and}
  \bibinfo{person}{Lei Zhang}.} \bibinfo{year}{2017}\natexlab{}.
\newblock \showarticletitle{Bottom-up and top-down attention for image
  captioning and VQA}.
\newblock \bibinfo{journal}{\emph{arXiv preprint arXiv:1707.07998}}
  (\bibinfo{year}{2017}).
\newblock


\bibitem[\protect\citeauthoryear{Antol, Agrawal, Lu, Mitchell, Batra,
  Lawrence~Zitnick, and Parikh}{Antol et~al\mbox{.}}{2015}]%
        {antol2015vqa}
\bibfield{author}{\bibinfo{person}{Stanislaw Antol}, \bibinfo{person}{Aishwarya
  Agrawal}, \bibinfo{person}{Jiasen Lu}, \bibinfo{person}{Margaret Mitchell},
  \bibinfo{person}{Dhruv Batra}, \bibinfo{person}{C Lawrence~Zitnick}, {and}
  \bibinfo{person}{Devi Parikh}.} \bibinfo{year}{2015}\natexlab{}.
\newblock \showarticletitle{Vqa: Visual question answering}. In
  \bibinfo{booktitle}{\emph{ICCV}}.
\newblock


\bibitem[\protect\citeauthoryear{Bahdanau, Cho, and Bengio}{Bahdanau
  et~al\mbox{.}}{2015}]%
        {bahdanau2014neural}
\bibfield{author}{\bibinfo{person}{Dzmitry Bahdanau},
  \bibinfo{person}{Kyunghyun Cho}, {and} \bibinfo{person}{Yoshua Bengio}.}
  \bibinfo{year}{2015}\natexlab{}.
\newblock \showarticletitle{Neural machine translation by jointly learning to
  align and translate}. In \bibinfo{booktitle}{\emph{ICLR}}.
\newblock


\bibitem[\protect\citeauthoryear{Banerjee and Lavie}{Banerjee and
  Lavie}{2005}]%
        {banerjee2005meteor}
\bibfield{author}{\bibinfo{person}{Satanjeev Banerjee} {and}
  \bibinfo{person}{Alon Lavie}.} \bibinfo{year}{2005}\natexlab{}.
\newblock \showarticletitle{METEOR: An automatic metric for MT evaluation with
  improved correlation with human judgments}. In
  \bibinfo{booktitle}{\emph{Proceedings of the acl workshop on intrinsic and
  extrinsic evaluation measures for machine translation and/or summarization}}.
\newblock


\bibitem[\protect\citeauthoryear{Chen, Ji, Su, Wu, and Wu}{Chen
  et~al\mbox{.}}{2017a}]%
        {chen2017structcap}
\bibfield{author}{\bibinfo{person}{Fuhai Chen}, \bibinfo{person}{Rongrong Ji},
  \bibinfo{person}{Jinsong Su}, \bibinfo{person}{Yongjian Wu}, {and}
  \bibinfo{person}{Yunsheng Wu}.} \bibinfo{year}{2017}\natexlab{a}.
\newblock \showarticletitle{Structcap: Structured semantic embedding for image
  captioning}. In \bibinfo{booktitle}{\emph{Proceedings of the 2017 ACM on
  Multimedia Conference}}. ACM, \bibinfo{pages}{46--54}.
\newblock


\bibitem[\protect\citeauthoryear{Chen, Zhang, Xiao, Nie, Shao, Liu, and
  Chua}{Chen et~al\mbox{.}}{2017b}]%
        {chen2017sca}
\bibfield{author}{\bibinfo{person}{Long Chen}, \bibinfo{person}{Hanwang Zhang},
  \bibinfo{person}{Jun Xiao}, \bibinfo{person}{Liqiang Nie},
  \bibinfo{person}{Jian Shao}, \bibinfo{person}{Wei Liu}, {and}
  \bibinfo{person}{Tat-Seng Chua}.} \bibinfo{year}{2017}\natexlab{b}.
\newblock \showarticletitle{Sca-cnn: Spatial and channel-wise attention in
  convolutional networks for image captioning}. In
  \bibinfo{booktitle}{\emph{CVPR}}.
\newblock


\bibitem[\protect\citeauthoryear{Das, Kottur, Gupta, Singh, Yadav, Moura,
  Parikh, and Batra}{Das et~al\mbox{.}}{2017}]%
        {das2017visual}
\bibfield{author}{\bibinfo{person}{Abhishek Das}, \bibinfo{person}{Satwik
  Kottur}, \bibinfo{person}{Khushi Gupta}, \bibinfo{person}{Avi Singh},
  \bibinfo{person}{Deshraj Yadav}, \bibinfo{person}{Jos{\'e}~MF Moura},
  \bibinfo{person}{Devi Parikh}, {and} \bibinfo{person}{Dhruv Batra}.}
  \bibinfo{year}{2017}\natexlab{}.
\newblock \showarticletitle{Visual dialog}. In
  \bibinfo{booktitle}{\emph{CVPR}}.
\newblock


\bibitem[\protect\citeauthoryear{Fang, Gupta, Iandola, Srivastava, Deng,
  Doll{\'a}r, Gao, He, Mitchell, Platt, et~al\mbox{.}}{Fang
  et~al\mbox{.}}{2015}]%
        {fang2015captions}
\bibfield{author}{\bibinfo{person}{Hao Fang}, \bibinfo{person}{Saurabh Gupta},
  \bibinfo{person}{Forrest Iandola}, \bibinfo{person}{Rupesh Srivastava},
  \bibinfo{person}{Li Deng}, \bibinfo{person}{Piotr Doll{\'a}r},
  \bibinfo{person}{Jianfeng Gao}, \bibinfo{person}{Xiaodong He},
  \bibinfo{person}{Margaret Mitchell}, \bibinfo{person}{John Platt},
  {et~al\mbox{.}}} \bibinfo{year}{2015}\natexlab{}.
\newblock \showarticletitle{From captions to visual concepts and back}.
\newblock  (\bibinfo{year}{2015}).
\newblock


\bibitem[\protect\citeauthoryear{Gu, Cai, Wang, and Chen}{Gu
  et~al\mbox{.}}{2017}]%
        {gu2017stack}
\bibfield{author}{\bibinfo{person}{Jiuxiang Gu}, \bibinfo{person}{Jianfei Cai},
  \bibinfo{person}{Gang Wang}, {and} \bibinfo{person}{Tsuhan Chen}.}
  \bibinfo{year}{2017}\natexlab{}.
\newblock \showarticletitle{Stack-captioning: Coarse-to-fine learning for image
  captioning}.
\newblock \bibinfo{journal}{\emph{arXiv preprint arXiv:1709.03376}}
  (\bibinfo{year}{2017}).
\newblock


\bibitem[\protect\citeauthoryear{He, Zhang, Ren, and Sun}{He
  et~al\mbox{.}}{2016}]%
        {he2016deep}
\bibfield{author}{\bibinfo{person}{Kaiming He}, \bibinfo{person}{Xiangyu
  Zhang}, \bibinfo{person}{Shaoqing Ren}, {and} \bibinfo{person}{Jian Sun}.}
  \bibinfo{year}{2016}\natexlab{}.
\newblock \showarticletitle{Deep residual learning for image recognition}. In
  \bibinfo{booktitle}{\emph{CVPR}}.
\newblock


\bibitem[\protect\citeauthoryear{Hochreiter and Schmidhuber}{Hochreiter and
  Schmidhuber}{1997}]%
        {hochreiter1997long}
\bibfield{author}{\bibinfo{person}{Sepp Hochreiter} {and}
  \bibinfo{person}{J{\"u}rgen Schmidhuber}.} \bibinfo{year}{1997}\natexlab{}.
\newblock \showarticletitle{Long short-term memory}.
\newblock \bibinfo{journal}{\emph{Neural computation}} (\bibinfo{year}{1997}).
\newblock


\bibitem[\protect\citeauthoryear{Hu, Andreas, Rohrbach, Darrell, and Saenko}{Hu
  et~al\mbox{.}}{2017}]%
        {hu2017learning}
\bibfield{author}{\bibinfo{person}{Ronghang Hu}, \bibinfo{person}{Jacob
  Andreas}, \bibinfo{person}{Marcus Rohrbach}, \bibinfo{person}{Trevor
  Darrell}, {and} \bibinfo{person}{Kate Saenko}.}
  \bibinfo{year}{2017}\natexlab{}.
\newblock \showarticletitle{Learning to Reason: End-To-End Module Networks for
  Visual Question Answering}. In \bibinfo{booktitle}{\emph{ICCV}}.
\newblock


\bibitem[\protect\citeauthoryear{Jabri, Joulin, and van~der Maaten}{Jabri
  et~al\mbox{.}}{2016}]%
        {jabri2016revisiting}
\bibfield{author}{\bibinfo{person}{Allan Jabri}, \bibinfo{person}{Armand
  Joulin}, {and} \bibinfo{person}{Laurens van~der Maaten}.}
  \bibinfo{year}{2016}\natexlab{}.
\newblock \showarticletitle{Revisiting visual question answering baselines}. In
  \bibinfo{booktitle}{\emph{ECCV}}.
\newblock


\bibitem[\protect\citeauthoryear{Johnson, Hariharan, van~der Maaten, Hoffman,
  Fei-Fei, Lawrence~Zitnick, and Girshick}{Johnson et~al\mbox{.}}{2017}]%
        {johnson2017inferring}
\bibfield{author}{\bibinfo{person}{Justin Johnson}, \bibinfo{person}{Bharath
  Hariharan}, \bibinfo{person}{Laurens van~der Maaten}, \bibinfo{person}{Judy
  Hoffman}, \bibinfo{person}{Li Fei-Fei}, \bibinfo{person}{C Lawrence~Zitnick},
  {and} \bibinfo{person}{Ross Girshick}.} \bibinfo{year}{2017}\natexlab{}.
\newblock \showarticletitle{Inferring and Executing Programs for Visual
  Reasoning}. In \bibinfo{booktitle}{\emph{ICCV}}.
\newblock


\bibitem[\protect\citeauthoryear{Karpathy and Fei-Fei}{Karpathy and
  Fei-Fei}{2015}]%
        {karpathy2015deep}
\bibfield{author}{\bibinfo{person}{Andrej Karpathy} {and} \bibinfo{person}{Li
  Fei-Fei}.} \bibinfo{year}{2015}\natexlab{}.
\newblock \showarticletitle{Deep visual-semantic alignments for generating
  image descriptions}. In \bibinfo{booktitle}{\emph{CVPR}}.
\newblock


\bibitem[\protect\citeauthoryear{Kingma and Ba}{Kingma and Ba}{2014}]%
        {kingma2014adam}
\bibfield{author}{\bibinfo{person}{Diederik~P Kingma} {and}
  \bibinfo{person}{Jimmy Ba}.} \bibinfo{year}{2014}\natexlab{}.
\newblock \showarticletitle{Adam: A method for stochastic optimization}.
\newblock \bibinfo{journal}{\emph{arXiv preprint arXiv:1412.6980}}
  (\bibinfo{year}{2014}).
\newblock


\bibitem[\protect\citeauthoryear{Krishna, Zhu, Groth, Johnson, Hata, Kravitz,
  Chen, Kalantidis, Li, Shamma, et~al\mbox{.}}{Krishna et~al\mbox{.}}{2017}]%
        {krishna2017visual}
\bibfield{author}{\bibinfo{person}{Ranjay Krishna}, \bibinfo{person}{Yuke Zhu},
  \bibinfo{person}{Oliver Groth}, \bibinfo{person}{Justin Johnson},
  \bibinfo{person}{Kenji Hata}, \bibinfo{person}{Joshua Kravitz},
  \bibinfo{person}{Stephanie Chen}, \bibinfo{person}{Yannis Kalantidis},
  \bibinfo{person}{Li-Jia Li}, \bibinfo{person}{David~A Shamma},
  {et~al\mbox{.}}} \bibinfo{year}{2017}\natexlab{}.
\newblock \showarticletitle{Visual genome: Connecting language and vision using
  crowdsourced dense image annotations}.
\newblock \bibinfo{journal}{\emph{IJCV}} (\bibinfo{year}{2017}).
\newblock


\bibitem[\protect\citeauthoryear{Krizhevsky, Sutskever, and Hinton}{Krizhevsky
  et~al\mbox{.}}{2012}]%
        {krizhevsky2012imagenet}
\bibfield{author}{\bibinfo{person}{Alex Krizhevsky}, \bibinfo{person}{Ilya
  Sutskever}, {and} \bibinfo{person}{Geoffrey~E Hinton}.}
  \bibinfo{year}{2012}\natexlab{}.
\newblock \showarticletitle{Imagenet classification with deep convolutional
  neural networks}. In \bibinfo{booktitle}{\emph{NIPS}}.
\newblock


\bibitem[\protect\citeauthoryear{Lake, Ullman, Tenenbaum, and Gershman}{Lake
  et~al\mbox{.}}{2017}]%
        {lake2017building}
\bibfield{author}{\bibinfo{person}{Brenden~M Lake}, \bibinfo{person}{Tomer~D
  Ullman}, \bibinfo{person}{Joshua~B Tenenbaum}, {and}
  \bibinfo{person}{Samuel~J Gershman}.} \bibinfo{year}{2017}\natexlab{}.
\newblock \showarticletitle{Building machines that learn and think like
  people}.
\newblock \bibinfo{journal}{\emph{Behavioral and Brain Sciences}}
  (\bibinfo{year}{2017}).
\newblock


\bibitem[\protect\citeauthoryear{Lin}{Lin}{2004}]%
        {lin2004rouge}
\bibfield{author}{\bibinfo{person}{Chin-Yew Lin}.}
  \bibinfo{year}{2004}\natexlab{}.
\newblock \showarticletitle{Rouge: A package for automatic evaluation of
  summaries}.
\newblock \bibinfo{journal}{\emph{Text Summarization Branches Out}}
  (\bibinfo{year}{2004}).
\newblock


\bibitem[\protect\citeauthoryear{Lin, Maire, Belongie, Hays, Perona, Ramanan,
  Doll{\'a}r, and Zitnick}{Lin et~al\mbox{.}}{2014}]%
        {lin2014microsoft}
\bibfield{author}{\bibinfo{person}{Tsung-Yi Lin}, \bibinfo{person}{Michael
  Maire}, \bibinfo{person}{Serge Belongie}, \bibinfo{person}{James Hays},
  \bibinfo{person}{Pietro Perona}, \bibinfo{person}{Deva Ramanan},
  \bibinfo{person}{Piotr Doll{\'a}r}, {and} \bibinfo{person}{C~Lawrence
  Zitnick}.} \bibinfo{year}{2014}\natexlab{}.
\newblock \showarticletitle{Microsoft coco: Common objects in context}. In
  \bibinfo{booktitle}{\emph{ECCV}}.
\newblock


\bibitem[\protect\citeauthoryear{Liu, Zhu, Ye, Guadarrama, and Murphy}{Liu
  et~al\mbox{.}}{2016}]%
        {liu2016improved}
\bibfield{author}{\bibinfo{person}{Siqi Liu}, \bibinfo{person}{Zhenhai Zhu},
  \bibinfo{person}{Ning Ye}, \bibinfo{person}{Sergio Guadarrama}, {and}
  \bibinfo{person}{Kevin Murphy}.} \bibinfo{year}{2016}\natexlab{}.
\newblock \showarticletitle{Improved image captioning via policy gradient
  optimization of spider}.
\newblock \bibinfo{journal}{\emph{arXiv preprint arXiv:1612.00370}}
  (\bibinfo{year}{2016}).
\newblock


\bibitem[\protect\citeauthoryear{Lu, Xiong, Parikh, and Socher}{Lu
  et~al\mbox{.}}{[n. d.]}]%
        {lu2017knowing}
\bibfield{author}{\bibinfo{person}{Jiasen Lu}, \bibinfo{person}{Caiming Xiong},
  \bibinfo{person}{Devi Parikh}, {and} \bibinfo{person}{Richard Socher}.}
  \bibinfo{year}{[n. d.]}\natexlab{}.
\newblock \showarticletitle{Knowing when to look: Adaptive attention via a
  visual sentinel for image captioning}. In \bibinfo{booktitle}{\emph{CVPR}}.
\newblock


\bibitem[\protect\citeauthoryear{Mao, Xu, Yang, Wang, Huang, and Yuille}{Mao
  et~al\mbox{.}}{2014}]%
        {mao2014deep}
\bibfield{author}{\bibinfo{person}{Junhua Mao}, \bibinfo{person}{Wei Xu},
  \bibinfo{person}{Yi Yang}, \bibinfo{person}{Jiang Wang},
  \bibinfo{person}{Zhiheng Huang}, {and} \bibinfo{person}{Alan Yuille}.}
  \bibinfo{year}{2014}\natexlab{}.
\newblock \showarticletitle{Deep captioning with multimodal recurrent neural
  networks (m-rnn)}.
\newblock \bibinfo{journal}{\emph{arXiv preprint arXiv:1412.6632}}
  (\bibinfo{year}{2014}).
\newblock


\bibitem[\protect\citeauthoryear{Papineni, Roukos, Ward, and Zhu}{Papineni
  et~al\mbox{.}}{2002}]%
        {papineni2002bleu}
\bibfield{author}{\bibinfo{person}{Kishore Papineni}, \bibinfo{person}{Salim
  Roukos}, \bibinfo{person}{Todd Ward}, {and} \bibinfo{person}{Wei-Jing Zhu}.}
  \bibinfo{year}{2002}\natexlab{}.
\newblock \showarticletitle{BLEU: a method for automatic evaluation of machine
  translation}. In \bibinfo{booktitle}{\emph{ACL}}.
\newblock


\bibitem[\protect\citeauthoryear{Ranzato, Chopra, Auli, and Zaremba}{Ranzato
  et~al\mbox{.}}{2015}]%
        {ranzato2015sequence}
\bibfield{author}{\bibinfo{person}{Marc'Aurelio Ranzato},
  \bibinfo{person}{Sumit Chopra}, \bibinfo{person}{Michael Auli}, {and}
  \bibinfo{person}{Wojciech Zaremba}.} \bibinfo{year}{2015}\natexlab{}.
\newblock \showarticletitle{Sequence level training with recurrent neural
  networks}.
\newblock \bibinfo{journal}{\emph{arXiv preprint arXiv:1511.06732}}
  (\bibinfo{year}{2015}).
\newblock


\bibitem[\protect\citeauthoryear{Ren, He, Girshick, and Sun}{Ren
  et~al\mbox{.}}{2015}]%
        {ren2015faster}
\bibfield{author}{\bibinfo{person}{Shaoqing Ren}, \bibinfo{person}{Kaiming He},
  \bibinfo{person}{Ross Girshick}, {and} \bibinfo{person}{Jian Sun}.}
  \bibinfo{year}{2015}\natexlab{}.
\newblock \showarticletitle{Faster r-cnn: Towards real-time object detection
  with region proposal networks}. In \bibinfo{booktitle}{\emph{NIPS}}.
\newblock


\bibitem[\protect\citeauthoryear{Ren, Wang, Zhang, Lv, and Li}{Ren
  et~al\mbox{.}}{2017}]%
        {ren2017deep}
\bibfield{author}{\bibinfo{person}{Zhou Ren}, \bibinfo{person}{Xiaoyu Wang},
  \bibinfo{person}{Ning Zhang}, \bibinfo{person}{Xutao Lv}, {and}
  \bibinfo{person}{Li-Jia Li}.} \bibinfo{year}{2017}\natexlab{}.
\newblock \showarticletitle{Deep reinforcement learning-based image captioning
  with embedding reward}.
\newblock \bibinfo{journal}{\emph{arXiv preprint arXiv:1704.03899}}
  (\bibinfo{year}{2017}).
\newblock


\bibitem[\protect\citeauthoryear{Rennie, Marcheret, Mroueh, Ross, and
  Goel}{Rennie et~al\mbox{.}}{2016}]%
        {rennie2016self}
\bibfield{author}{\bibinfo{person}{Steven~J Rennie}, \bibinfo{person}{Etienne
  Marcheret}, \bibinfo{person}{Youssef Mroueh}, \bibinfo{person}{Jarret Ross},
  {and} \bibinfo{person}{Vaibhava Goel}.} \bibinfo{year}{2016}\natexlab{}.
\newblock \showarticletitle{Self-critical sequence training for image
  captioning}.
\newblock \bibinfo{journal}{\emph{arXiv preprint arXiv:1612.00563}}
  (\bibinfo{year}{2016}).
\newblock


\bibitem[\protect\citeauthoryear{Russakovsky, Deng, Su, Krause, Satheesh, Ma,
  Huang, Karpathy, Khosla, Bernstein, et~al\mbox{.}}{Russakovsky
  et~al\mbox{.}}{2015}]%
        {russakovsky2015imagenet}
\bibfield{author}{\bibinfo{person}{Olga Russakovsky}, \bibinfo{person}{Jia
  Deng}, \bibinfo{person}{Hao Su}, \bibinfo{person}{Jonathan Krause},
  \bibinfo{person}{Sanjeev Satheesh}, \bibinfo{person}{Sean Ma},
  \bibinfo{person}{Zhiheng Huang}, \bibinfo{person}{Andrej Karpathy},
  \bibinfo{person}{Aditya Khosla}, \bibinfo{person}{Michael Bernstein},
  {et~al\mbox{.}}} \bibinfo{year}{2015}\natexlab{}.
\newblock \showarticletitle{Imagenet large scale visual recognition challenge}.
\newblock \bibinfo{journal}{\emph{IJCV}} (\bibinfo{year}{2015}).
\newblock


\bibitem[\protect\citeauthoryear{Stanfill and Waltz}{Stanfill and
  Waltz}{1986}]%
        {stanfill1986toward}
\bibfield{author}{\bibinfo{person}{Craig Stanfill} {and} \bibinfo{person}{David
  Waltz}.} \bibinfo{year}{1986}\natexlab{}.
\newblock \showarticletitle{Toward memory-based reasoning}.
\newblock \bibinfo{journal}{\emph{Commun. ACM}} (\bibinfo{year}{1986}).
\newblock


\bibitem[\protect\citeauthoryear{Sutton and Barto}{Sutton and Barto}{1998}]%
        {sutton1998reinforcement}
\bibfield{author}{\bibinfo{person}{Richard~S Sutton} {and}
  \bibinfo{person}{Andrew~G Barto}.} \bibinfo{year}{1998}\natexlab{}.
\newblock \bibinfo{booktitle}{\emph{Reinforcement learning: An introduction}}.
\newblock \bibinfo{publisher}{MIT press Cambridge}.
\newblock


\bibitem[\protect\citeauthoryear{Toutanova, Klein, Manning, and
  Singer}{Toutanova et~al\mbox{.}}{2003}]%
        {toutanova2003feature}
\bibfield{author}{\bibinfo{person}{Kristina Toutanova}, \bibinfo{person}{Dan
  Klein}, \bibinfo{person}{Christopher~D Manning}, {and} \bibinfo{person}{Yoram
  Singer}.} \bibinfo{year}{2003}\natexlab{}.
\newblock \showarticletitle{Feature-rich part-of-speech tagging with a cyclic
  dependency network}. In \bibinfo{booktitle}{\emph{Proceedings of the 2003
  Conference of the North American Chapter of the Association for Computational
  Linguistics on Human Language Technology-Volume 1}}.
\newblock


\bibitem[\protect\citeauthoryear{Vedantam, Lawrence~Zitnick, and
  Parikh}{Vedantam et~al\mbox{.}}{2015}]%
        {vedantam2015cider}
\bibfield{author}{\bibinfo{person}{Ramakrishna Vedantam}, \bibinfo{person}{C
  Lawrence~Zitnick}, {and} \bibinfo{person}{Devi Parikh}.}
  \bibinfo{year}{2015}\natexlab{}.
\newblock \showarticletitle{Cider: Consensus-based image description
  evaluation}. In \bibinfo{booktitle}{\emph{CVPR}}.
\newblock


\bibitem[\protect\citeauthoryear{Vinyals, Toshev, Bengio, and Erhan}{Vinyals
  et~al\mbox{.}}{2015}]%
        {vinyals2015show}
\bibfield{author}{\bibinfo{person}{Oriol Vinyals}, \bibinfo{person}{Alexander
  Toshev}, \bibinfo{person}{Samy Bengio}, {and} \bibinfo{person}{Dumitru
  Erhan}.} \bibinfo{year}{2015}\natexlab{}.
\newblock \showarticletitle{Show and tell: A neural image caption generator}.
  In \bibinfo{booktitle}{\emph{CVPR}}.
\newblock


\bibitem[\protect\citeauthoryear{Williams}{Williams}{1992}]%
        {williams1992simple}
\bibfield{author}{\bibinfo{person}{Ronald~J Williams}.}
  \bibinfo{year}{1992}\natexlab{}.
\newblock \showarticletitle{Simple statistical gradient-following algorithms
  for connectionist reinforcement learning}.
\newblock In \bibinfo{booktitle}{\emph{Reinforcement Learning}}.
\newblock


\bibitem[\protect\citeauthoryear{Xu, Zhu, Choy, and Fei-Fei}{Xu
  et~al\mbox{.}}{2017}]%
        {xu2017scene}
\bibfield{author}{\bibinfo{person}{Danfei Xu}, \bibinfo{person}{Yuke Zhu},
  \bibinfo{person}{Christopher~B Choy}, {and} \bibinfo{person}{Li Fei-Fei}.}
  \bibinfo{year}{2017}\natexlab{}.
\newblock \showarticletitle{Scene graph generation by iterative message
  passing}. In \bibinfo{booktitle}{\emph{CVPR}}.
\newblock


\bibitem[\protect\citeauthoryear{Xu, Ba, Kiros, Cho, Courville, Salakhudinov,
  Zemel, and Bengio}{Xu et~al\mbox{.}}{2015}]%
        {xu2015show}
\bibfield{author}{\bibinfo{person}{Kelvin Xu}, \bibinfo{person}{Jimmy Ba},
  \bibinfo{person}{Ryan Kiros}, \bibinfo{person}{Kyunghyun Cho},
  \bibinfo{person}{Aaron Courville}, \bibinfo{person}{Ruslan Salakhudinov},
  \bibinfo{person}{Rich Zemel}, {and} \bibinfo{person}{Yoshua Bengio}.}
  \bibinfo{year}{2015}\natexlab{}.
\newblock \showarticletitle{Show, attend and tell: Neural image caption
  generation with visual attention}. In \bibinfo{booktitle}{\emph{ICML}}.
\newblock


\bibitem[\protect\citeauthoryear{Yao, Pan, Li, Qiu, and Mei}{Yao
  et~al\mbox{.}}{2016}]%
        {yao2016boosting}
\bibfield{author}{\bibinfo{person}{Ting Yao}, \bibinfo{person}{Yingwei Pan},
  \bibinfo{person}{Yehao Li}, \bibinfo{person}{Zhaofan Qiu}, {and}
  \bibinfo{person}{Tao Mei}.} \bibinfo{year}{2016}\natexlab{}.
\newblock \showarticletitle{Boosting image captioning with attributes}.
\newblock \bibinfo{journal}{\emph{OpenReview}} (\bibinfo{year}{2016}).
\newblock


\bibitem[\protect\citeauthoryear{Yu, Zhang, Wang, and Yu}{Yu
  et~al\mbox{.}}{2017}]%
        {yu2017seqgan}
\bibfield{author}{\bibinfo{person}{Lantao Yu}, \bibinfo{person}{Weinan Zhang},
  \bibinfo{person}{Jun Wang}, {and} \bibinfo{person}{Yong Yu}.}
  \bibinfo{year}{2017}\natexlab{}.
\newblock \showarticletitle{SeqGAN: Sequence Generative Adversarial Nets with
  Policy Gradient.}. In \bibinfo{booktitle}{\emph{AAAI}}.
\newblock


\bibitem[\protect\citeauthoryear{Zhang, Kyaw, Chang, and Chua}{Zhang
  et~al\mbox{.}}{2017a}]%
        {zhang2017visual}
\bibfield{author}{\bibinfo{person}{Hanwang Zhang}, \bibinfo{person}{Zawlin
  Kyaw}, \bibinfo{person}{Shih-Fu Chang}, {and} \bibinfo{person}{Tat-Seng
  Chua}.} \bibinfo{year}{2017}\natexlab{a}.
\newblock \showarticletitle{Visual translation embedding network for visual
  relation detection}. In \bibinfo{booktitle}{\emph{CVPR}}.
\newblock


\bibitem[\protect\citeauthoryear{Zhang, Niu, and Chang}{Zhang
  et~al\mbox{.}}{2018}]%
        {zhang2018grounding}
\bibfield{author}{\bibinfo{person}{Hanwang Zhang}, \bibinfo{person}{Yulei Niu},
  {and} \bibinfo{person}{Shih-Fu Chang}.} \bibinfo{year}{2018}\natexlab{}.
\newblock \showarticletitle{Grounding Referring Expressions in Images by
  Variational Context}. In \bibinfo{booktitle}{\emph{CVPR}}.
\newblock


\bibitem[\protect\citeauthoryear{Zhang, Sung, Liu, Xiang, Gong, Yang, and
  Hospedales}{Zhang et~al\mbox{.}}{2017b}]%
        {zhang2017actor}
\bibfield{author}{\bibinfo{person}{Li Zhang}, \bibinfo{person}{Flood Sung},
  \bibinfo{person}{Feng Liu}, \bibinfo{person}{Tao Xiang},
  \bibinfo{person}{Shaogang Gong}, \bibinfo{person}{Yongxin Yang}, {and}
  \bibinfo{person}{Timothy~M Hospedales}.} \bibinfo{year}{2017}\natexlab{b}.
\newblock \showarticletitle{Actor-Critic Sequence Training for Image
  Captioning}.
\newblock \bibinfo{journal}{\emph{arXiv preprint arXiv:1706.09601}}
  (\bibinfo{year}{2017}).
\newblock


\end{thebibliography}

\end{document}